\documentclass[]{imag-ms-template}

\usepackage{longtable} 

\usepackage[utf8]{inputenc} 
\usepackage[T1]{fontenc}  
\usepackage{csquotes}    
\usepackage[backend=biber, style=apa]{biblatex}

\usepackage{placeins}
\usepackage{caption}

\title{Wavelet-Fusion Diffusion Model for Multimodal Brain MRI Synthesis with Modality and Metadata Conditioning}

\author{Muhammad Nabi Yasinzai,$^{1}$ Remika Mito,$^{2}$ Mangor Pedersen$^{1\ast}$\\
{\small $^{1}$Department of Psychology \& Neuroscience, Auckland University of Technology,}\\
{\small Auckland, New Zealand.}\\
{\small $^{2}$Department of Psychiatry, University of Melbourne, Melbourne, Victoria 3053, Australia.}\\
{\small $^\ast$Correspondence: mangor.pedersen@aut.ac.nz}
}

\begin{document}

\maketitle 


\begin{abstract}
Multimodal MRI provides complementary information for neuroimaging analysis, where different imaging modalities capture distinct anatomical, tissue, and pathological features that support the development and evaluation of downstream artificial intelligence (AI) applications.
Although large-scale structural MRI resources are increasingly available, their modality coverage is often uneven across public and pooled neuroimaging datasets. 
T1-weighted (T1w) MRI is widely represented, whereas T2-weighted (T2w) and fluid-attenuated inversion recovery (FLAIR) images are less consistently available.
Gadolinium-enhanced T1-weighted imaging (T1Gd) is particularly scarce in public neuroimaging datasets because it is mainly acquired when clinically indicated, including for brain tumor assessment.
This uneven modality coverage is further complicated by heterogeneity across sites, scanners, and acquisition protocols, as well as demographic and clinical variables that are often sparse, inconsistently recorded, or unavailable across studies.
Synthetic MRI generation can help address this imbalance by synthesizing target-modality volumes for dataset augmentation and controlled synthetic cohort creation.
However, many existing MRI synthesis approaches are trained on narrow modality sets or relatively homogeneous cohorts, limiting their applicability to large pooled neuroimaging resources where modality availability, acquisition protocols, and metadata coverage vary substantially across datasets.
Diffusion models have become an attractive approach for MRI synthesis because of their strong sample fidelity and diversity, but sampling directly in 3D voxel space is computationally expensive and slow at inference.
Latent diffusion improves practicality by synthesizing MRI in a learned, compressed 3D latent space, although generation quality depends on the autoencoder's reconstruction fidelity and the resulting latent distribution.
Building on BrainScape, our previously developed open-source framework and curated resource for large-scale aggregation and preprocessing of anatomical MRI, we develop a 3D latent diffusion model for controllable multimodal brain MRI synthesis conditioned on the target modality and demographic/clinical metadata.
For model training and evaluation, we use a fixed earlier BrainScape snapshot comprising 157 datasets and 45{,}880 preprocessed MRI scans across T1w, T2w, T1Gd, and FLAIR, with demographic and, where available, clinical metadata. 
Our approach combines a Wavelet-Fusion variational autoencoder (WF-VAE) latent compressor with a conditional 3D U-Net diffusion model trained in the learned latent space using explicit modality and metadata conditioning.
In pooled BrainScape evaluation, our proposed Wavelet-Fusion Diffusion Model (WFDM) achieved 
the strongest distributional alignment among the evaluated synthetic MRI generators. 
Specifically, WFDM reduced pooled MedicalNet feature-space Fr\'echet distance (FID) from 0.00550 to 0.00404 
and pooled Maximum Mean Discrepancy (MMD) from 0.1918 to 0.1471 relative to the MAISI baseline trained on BrainScape, 
while maintaining comparable synthetic-sample diversity measured using multi-scale structural similarity (MS-SSIM). 
Modality-specific analysis showed improvements for T1w and T2w synthesis and smaller improvements for FLAIR, whereas T1Gd remained the most challenging contrast, with the Medical AI for Synthetic Imaging (MAISI) baseline achieving lower FID and MMD for this modality.
A sampler study further showed that Rectified Flow (RFlow) inference with 50 sampling steps achieved strong distributional realism at substantially lower sampling cost than 1000-step Denoising Diffusion Probabilistic Model (DDPM) variants. 
We also publicly release \textit{medmetric}, a Python-based package for standardized evaluation of 3D medical image synthesis models, 
providing metrics for assessing synthetic image realism and diversity. 
Overall, by learning from heterogeneous pooled neuroimaging data, WFDM provides a controllable framework for synthesizing underrepresented MRI modalities and generating synthetic cohorts with specified imaging and metadata characteristics. This supports more controlled scientific evaluation of how modality availability, cohort composition, and acquisition heterogeneity influence multimodal AI performance, while providing a clinically relevant basis for developing and evaluating AI methods when modality coverage is incomplete or uneven.
\end{abstract}

\FloatBarrier
\section{Introduction}

Multimodal structural MRI provides complementary information for neuroimaging analysis and clinical artificial intelligence (AI), because different MRI modalities capture distinct anatomical structures, tissue properties, and pathological features. 
For example, in clinical neuroradiology, multimodal MRI is particularly important for the diagnosis, treatment planning, and monitoring of brain tumors, where T1-weighted (T1w), T2-weighted (T2w), gadolinium-enhanced T1-weighted (T1Gd), and fluid-attenuated inversion recovery (FLAIR) MRI provide complementary information for identifying and delineating tumor subregions \parencite{Menze-BRATS-2014,Bakas-BRATS-2018}. 
This dependence on multiple MRI contrasts makes modality availability an important consideration when developing and evaluating downstream AI models, because missing modalities can remove complementary anatomical and pathological information and reduce the reliability of downstream neuroimaging analyses \parencite{Kim-ALDM-2024,Meng-FLAIRSynth-2024}.

Although complete multimodal MRI protocols can be clinically justified for selected patient groups, acquiring the same set of modalities for every subject is often neither practical nor clinically appropriate. 
This is especially important in research cohorts and healthy controls, as MRI acquisition is shaped by diagnostic objectives, clinical feasibility, scan time, cost, patient tolerance, and safety considerations \parencite{Jiang-CoLaDiff-2023,Kim-ALDM-2024}. 
This acquisition constraint is most evident for T1Gd, which highlights abnormal contrast enhancement associated with blood-brain barrier disruption in clinically relevant pathology, including brain tumors \parencite{Zhou-GadoliniumCancerImaging-2012}. 
Because gadolinium-based contrast agents are generally used only when clinically justified, T1Gd is common in selected clinical imaging protocols but is rarely acquired in healthy or population cohorts. 
Consequently, neuroimaging datasets often have incomplete modality coverage, with many subjects lacking a complete set of multimodal MRI sequences.

Pooling datasets across studies, sites, and cohorts increases the statistical robustness, broadens demographic and clinical representation, and exposes downstream AI models to acquisition variability that is not captured by a single homogeneous dataset \parencite{Yasinzai-BrainScape-2025}. 
However, pooling MRI data does not inherently resolve the issue of missing modalities and can also introduce differences in scanner hardware, acquisition protocols, and preprocessing workflows that affect the reliability of downstream neuroimaging analyses \parencite{Dadar-MultisiteNeuroimaging-2020,Yasinzai-BrainScape-2025}.
Synthetic MRI generation can help address modality imbalance by synthesizing underrepresented MRI modalities for dataset augmentation and controlled synthetic cohort creation \parencite{Koetzier-SyntheticDataReview-2024}.
For datasets with incomplete multimodal MRI coverage, generative models can synthesize missing contrasts or translate between available modalities, supporting downstream pipelines that expect complete multimodal inputs \parencite{Jiang-CoLaDiff-2023,Kim-ALDM-2024,Meng-FLAIRSynth-2024}.
More broadly, synthetic medical images can enrich training data when real data are limited \parencite{Khader-DDPM3D-2023,Koetzier-SyntheticDataReview-2024}.
Metadata-conditioned brain MRI synthesis further enables generation under predefined cohort attributes, such as age and sex, supporting cohort-controlled generation and subgroup-aware model development \parencite{Peng-BrainSynth-2024}.
However, applying metadata conditioning to large pooled datasets requires handling incomplete metadata coverage, because demographic and clinical variables are often sparse, inconsistently recorded, or unavailable across studies \parencite{Yasinzai-BrainScape-2025}.

This pattern of modality imbalance, acquisition heterogeneity, and sparse metadata is directly reflected in BrainScape, an open-source framework that we previously developed to aggregate and preprocess public anatomical MRI datasets \parencite{Yasinzai-BrainScape-2025}.
BrainScape includes T1w, T2w, T1Gd, and FLAIR MRI across a large collection of public datasets, while preserving source provenance, acquisition information, and available demographic or clinical metadata \parencite{Yasinzai-BrainScape-2025}. 
This makes BrainScape useful for large-scale multimodal MRI synthesis, but also challenging because modality coverage, scanner hardware, acquisition protocols, and metadata availability vary substantially across source datasets. 
Prior work shows that text or metadata conditioning can guide brain MRI synthesis toward specified subject or cohort characteristics \parencite{Peng-BrainSynth-2024,Wang-TUMSyn2025-2024}.
In BrainScape, however, demographic and clinical metadata are not uniformly available across source datasets, so conditioning designs must use an explicit metadata availability mask that distinguishes observed covariate values from unavailable fields \parencite{Yasinzai-BrainScape-2025}.
This combination makes BrainScape a clinically relevant setting for controllable multimodal MRI synthesis under modality imbalance, acquisition heterogeneity, and sparse metadata.

Diffusion models have become an attractive approach for medical image synthesis because of their strong sample fidelity and diversity, but direct 3D voxel-space diffusion is computationally expensive and slow at inference \parencite{Kazerouni-MedDiffusionSurvey-2023,Fan-MRISurvey-2024}. 
To reduce this burden, previous work has used patch-based, slice-based, cascaded, or latent-space generation strategies \parencite{Bieder-MedDiffusionEfficiency-2024,Friedrich-WDM-2024,Zhu-MakeAVolume-2023}.
Latent diffusion addresses this challenge by learning a compact latent representation with an autoencoder and performing diffusion in the compressed latent space, reducing memory and compute requirements for high-resolution 3D MRI synthesis \parencite{Pinaya-LDM3DBrain-2022,Guo-Maisi-2025,Kim-ALDM-2024,Zhu-MakeAVolume-2023}. 
However, synthesis fidelity depends on the learned latent representation, because compression must reduce dimensionality while preserving fine anatomical structure, and information lost or distorted by the autoencoder can limit downstream diffusion quality \parencite{rombach-LDM-2022,Friedrich-WDM-2024}.
Similarly, wavelet-domain diffusion uses the discrete wavelet transform (DWT) to decompose a 3D volume into low-frequency and high-frequency subbands, enabling synthesis in a reduced spatial representation while retaining fine structural detail. 
The final image is then reconstructed from the generated subbands using the inverse discrete wavelet transform (IDWT).
This motivates latent compression designs that incorporate wavelet features to preserve fine structural information for volumetric medical image synthesis \parencite{Friedrich-WDM-2024}.
In contrast, patch-based or slice-based MRI generation strategies can reduce computational cost, but they may lose global 3D context, introduce slice or volumetric inconsistencies, or produce artifacts that limit downstream applicability \parencite{Friedrich-WDM-2024,Zhu-MakeAVolume-2023,Kim-ALDM-2024}.
For multimodal synthesis, diffusion models also benefit from explicit conditioning mechanisms, including source or target modality control and text or metadata conditioning, to guide generation toward the requested MRI modality or metadata context \parencite{Jiang-CoLaDiff-2023,Kim-ALDM-2024,Peng-BrainSynth-2024}.

Although diffusion models for synthetic MRI have progressed rapidly, many pipelines are still trained or evaluated within relatively narrow modality sets, homogeneous cohorts, or predefined contrast mappings, leaving open how well they generalize under cross-study heterogeneity \parencite{Pinaya-LDM3DBrain-2022,Jiang-CoLaDiff-2023,Zhang-MisaLDM-2025}.
General 3D latent diffusion frameworks developed for broad medical imaging provide strong scalable baselines, but they do not typically target the BrainScape setting, where a single generator must learn from heterogeneous pooled neuroimaging data with uneven modality availability, acquisition heterogeneity, and incomplete demographic or clinical metadata \parencite{Guo-Maisi-2025,Zhao-MaisiV2-2025,Yasinzai-BrainScape-2025}.
Although arbitrary missing-modality and multimodal translation approaches exist, robust unified generators that remain reliable across variable source and target modality availability, incomplete modality sets, and cross-dataset heterogeneity are still rarely demonstrated at large scale \parencite{Meng-FLAIRSynth-2024,Kim-ALDM-2024,Jiang-CoLaDiff-2023,Zhang-MisaLDM-2025}. 
Moreover, because modality coverage differs sharply across cohorts, models trained around a fixed modality configuration may have limited applicability to pooled datasets, motivating approaches that explicitly handle variable modality sets and missing modality inputs \parencite{Yasinzai-BrainScape-2025,Lupke-PhysInfBrainMRI-2024}.
Existing 3D medical diffusion and MRI translation pipelines often support controllability through modality conditioning, structural guidance, or spatial control signals such as anatomical masks, segmentation maps, and voxel spacing \parencite{Jiang-CoLaDiff-2023,Guo-Maisi-2025,Zhao-MaisiV2-2025}.
However, they do not directly address metadata conditioning in pooled neuroimaging datasets where demographic and clinical variables are sparse, inconsistently recorded, or absent across source datasets. 
Reliable metadata conditioning in this setting therefore requires an explicit metadata availability mask that distinguishes observed covariate values from unavailable fields \parencite{Yasinzai-BrainScape-2025}.

Finally, evaluation practices remain inconsistent across the literature, with studies differing in metric choice, preprocessing, normalization, feature representations, and whether evaluation is performed using 2D slice-based, 2.5D, or 3D protocols. 
Because these choices can affect assessments of realism, diversity, anatomical plausibility, and downstream utility, direct comparison and reproducibility across synthesis methods remain difficult \parencite{Koetzier-SyntheticDataReview-2024,Dohmen-SynthEvalMetrics-2025}. 

Building on these gaps, we develop a 3D latent diffusion pipeline for multimodal brain MRI synthesis designed for heterogeneous, pooled neuroimaging collections, supporting controllable generation across MRI modalities (T1w/T2w/FLAIR/T1Gd) using explicit target-modality conditioning and cohort metadata conditioning when available \parencite{Yasinzai-BrainScape-2025}. 
Our approach combines a Wavelet-Fusion VAE compression stage with a conditional 3D U-Net diffusion model trained in the learned latent space, where generation is guided by the target modality and by demographic and clinical metadata, with an explicit metadata availability mask.
We implement explicit target-modality conditioning to generate specific MRI contrasts within a single modality-conditioned generative model.
We extend the diffusion model with demographic and clinical metadata conditioning tailored to BrainScape's available metadata, enabling cohort-controlled synthesis and future subgroup analyses \parencite{Yasinzai-BrainScape-2025}. 
We report systematic experiments across compression choices, generator baselines, MRI modalities, and sampler configurations to quantify tradeoffs in fidelity, diversity, and inference cost for BrainScape multimodal MRI synthesis.


\FloatBarrier
\section{Methodology}

\subsection{Ethics statement}

This study is a secondary analysis of de-identified anatomical MRI data obtained from publicly available source datasets integrated through BrainScape. 
No new human data were collected for the present work. 
Ethical approval, participant consent, and requirements for data access are governed by the original contributing studies and repositories from which the BrainScape datasets were derived.

\subsection{Dataset}
\label{methodology-dataset}

\subsubsection{BrainScape overview}

The models are trained and evaluated on a fixed earlier BrainScape snapshot. The fully published BrainScape resource is a large-scale aggregation of 160 publicly available anatomical MRI datasets comprising 27{,}227 subjects and 46{,}583 scans after quality control.
BrainScape provides four anatomical MRI contrasts used in this study, including T1w, T2w, T1Gd, and FLAIR, and retains demographic, clinical, and acquisition metadata whenever these fields are available in the originating datasets \parencite{Yasinzai-BrainScape-2025}.

For the present study, we used a fixed BrainScape snapshot comprising 157 datasets and 45{,}880 preprocessed MRI scans across T1w, T2w, T1Gd, and FLAIR.
This snapshot was finalized before three additional datasets were incorporated into the published BrainScape release.
Consequently, the modality counts reported in Table~\ref{tab:brainscape-splits} differ slightly from the published BrainScape statistics.
The experimental snapshot contains fewer T1w and T2w scans, whereas the T1Gd and FLAIR counts are unchanged relative to the published BrainScape release.

Table~\ref{tab:brainscape-splits} summarizes the number of preprocessed BrainScape MRI volumes in the training, validation, and test splits used in this study.
We enforce subject-aware splitting so that no subject, across any modality or session, appears in more than one split, which prevents leakage and supports an unbiased evaluation.

\begin{table}[t]
\centering
\begin{threeparttable}
\caption{BrainScape split summary.\tnote{$a$}}
\label{tab:brainscape-splits}
\setlength{\tabcolsep}{30pt}%
\renewcommand{\arraystretch}{1.1}
\begin{tabular}{@{}lrrrr@{}}
\toprule
Modality & Total & Train & Val & Test \\
\midrule
T1w   & 30,965 & 24,515 & 3,036 & 3,414 \\
T2w   & 6,893  & 5,332  & 694   & 867   \\
FLAIR & 6,552  & 5,064  & 673   & 815   \\
T1Gd  & 1,470  & 1,001  & 172   & 297   \\
\midrule
Total & 45,880 & 35,912 & 4,575 & 5,393 \\
\bottomrule
\end{tabular}
\begin{tablenotes}[flushleft]\footnotesize
\item[${a}$]
Counts correspond to the fixed earlier BrainScape snapshot used for WFDM training, validation, and testing in this study.
This snapshot comprises 157 datasets and 45{,}880 MRI scans and was finalized before three additional datasets were incorporated into the published BrainScape release.
The published BrainScape release contains 160 datasets and 46{,}583 MRI scans after quality control, including 31{,}411 T1w, 7{,}150 T2w, 1{,}470 T1Gd, and 6{,}552 FLAIR scans \parencite{Yasinzai-BrainScape-2025}.
The difference is limited to T1w and T2w scans, whereas T1Gd and FLAIR counts are unchanged.
All experiments in this paper use the snapshot and subject-aware splits reported in this table.
\end{tablenotes}
\end{threeparttable}
\end{table}

\subsubsection{BrainScape metadata conditioning}
\label{methodology-metadata-conditioning}

Each BrainScape entry may also include different types of metadata, including continuous, categorical, and binary demographic variables and clinical indicators.
We use the available BrainScape demographic and clinical metadata for conditioning generative models.
The metadata conditioning uses a schema that defines a fixed order for 25 BrainScape metadata fields:
age, height, weight, BMI, age group, sex, race, handedness, education, socioeconomic status,
and a set of pathology indicators including stroke, schizophrenia, depression, Attention-Deficit/Hyperactivity Disorder (ADHD), bipolar disorder, prosopagnosia, epilepsy, Focal Cortical Dysplasia (FCD), Hippocampal Sclerosis (HS), tumor, acute ischaemic stroke, Dysembryoplastic Neuroepithelial Tumor (DNT), Gliosis (GL), aneurysm, and Autism Spectrum Disorder (ASD).
We convert the raw metadata for entry $i$ into an encoded vector of fixed length $d_i \in \mathbb{R}^{P}$ together with a metadata availability mask $r_i \in \{0,1\}^{P}$, where $P$ denotes the number of metadata fields listed in the schema and $(r_i)_j=1$ indicates that the $j$th metadata field is available for entry $i$.

Formally, the encoded metadata vector is defined as
\[
d_i = \mathrm{Enc}(\text{meta}_i)\in\mathbb{R}^{P},
\]
and the corresponding metadata availability mask is defined as
\[
 r_i \in \{0,1\}^{P}.
\]

Continuous variables, including age, height, weight, and BMI, are z-score normalized using BrainScape dataset statistics.
Categorical variables are mapped to integer identifiers, while binary demographic and clinical indicators are encoded as three-state \textit{yes}/\textit{no}/\textit{missing} integer labels, ensuring that an unrecorded value remains distinguishable from an explicit negative value.
The resulting feature vector $d_i$, together with the corresponding metadata availability mask $r_i$, is provided as input to a learnable metadata encoder, which produces the conditioning representation used by the underlying generative models.
This design enables the model to incorporate available demographic and clinical context while explicitly indicating which metadata fields are available.

\subsection{Preprocessing}

We preprocess the heterogeneous source datasets using the BrainScape BraTS preprocessing pipeline prior to model training \parencite{Yasinzai-BrainScape-2025}.
In brief, the pipeline aligns modalities within each subject, standardizes volumes to atlas space, performs brain extraction and intensity normalization, and resamples all volumes to a uniform resolution.

When multiple modalities are available for a given entry, intra-subject co-registration is performed to a single reference modality selected by the following priority order:
$\mathrm{T1w} > \mathrm{T2w} > \mathrm{T1Gd} > \mathrm{FLAIR}$.
This step improves spatial correspondence among available modalities within each subject and reduces contrast misalignment before model training.
Following co-registration, all volumes are rigidly registered to the SRI-24 atlas to standardize orientation and approximate anatomical scale \parencite{Yasinzai-BrainScape-2025}.
Brain extraction is then performed using HD-BET to generate a brain mask, which is used to remove non-brain regions from each MRI volume.
After registration and masking, the BrainScape BraTS pipeline applies intensity normalization and resamples all volumes to isotropic $1\,\mathrm{mm}$ spacing \parencite{Yasinzai-BrainScape-2025}.

For model training, volumes are represented in RAS orientation as channel-first tensors and cropped or padded to a fixed spatial size of $(H,W,D)=(192,224,160)$, giving tensor shape $(C,H,W,D)=(1,192,224,160)$.
In the PyTorch data module, each volume is additionally normalized using min-max scaling with high-percentile clipping before intensities are mapped to the range $[0,1]$.

\subsection{Problem formulation}

Let $\mathcal{M}$ denote the set of MRI contrasts considered in this study:
\begin{equation}
\mathcal{M}=\{\mathrm{T1w},\mathrm{T2w},\mathrm{T1Gd},\mathrm{FLAIR}\}, \qquad |\mathcal{M}|=4.
\label{eq:modality-set}
\end{equation}

For dataset subject entry $i$ and modality $m \in \mathcal{M}$, we write the corresponding preprocessed single-channel 3D volume as
\[
x_i^{(m)} \in \mathbb{R}^{H\times W\times D}, \qquad (H,W,D)=(192,224,160),
\]
Because not every subject entry contains every modality, let $\mathcal{M}_i^{\mathrm{avail}} \subseteq \mathcal{M}$ denote the set of modalities available for subject entry $i$, and define the corresponding set of observed volumes as
\[
X_i^{\mathrm{avail}} \triangleq \{x_i^{(m)} : m \in \mathcal{M}_i^{\mathrm{avail}}\}.
\]

For training, each training example is indexed by a subject entry $i$ and a selected target contrast $\tau$, where $\tau \in \mathcal{M}_i^{\mathrm{avail}}$. The symbol $\tau$ denotes the requested output MRI contrast and $x_i^{(\tau)}$ corresponds to the target MRI volume. The label $\tau$ is provided to the model as conditioning, and the corresponding volume serves as the output MRI target ($y_{i,\tau}$):
\begin{equation}
y_{i,\tau} \triangleq x_i^{(\tau)}.
\label{eq:target-volume}
\end{equation}

Using the metadata representation and metadata availability mask introduced in Section~\ref{methodology-metadata-conditioning}, the conditioning information associated with subject entry $i$ and target contrast $\tau$ is written as
\begin{equation}
c_{i,\tau} = (\tau, d_i, r_i),
\label{eq:conditioning-object}
\end{equation}
where $\tau$ specifies the desired output contrast, $d_i$ denotes the metadata feature vector, and $r_i$ is the corresponding metadata availability mask.
The objective is to train a conditional generative model that learns the distribution of target MRI volumes given the conditioning information $c_{i,\tau}$.

\subsection{Wavelet-Fusion Diffusion Model}
\label{sec:methodology-wfdm}

The proposed Wavelet-Fusion Diffusion Model (WFDM) is a 3D latent diffusion framework conditioned on modality and metadata for generating a target MRI contrast volume $y_{i,\tau}$ in a compact latent space rather than directly in voxel space.
Training is performed in two stages.
In Stage~A, a Wavelet-Fusion VAE compressor, denoted by the encoder and decoder pair $(E_{\phi},D_{\psi})$, learns a compact latent representation for preprocessed brain MRI volumes.
In Stage~B, a conditional diffusion model is trained on latent representations produced by the frozen selected Wavelet-Fusion VAE encoder and conditioned on the target modality, the metadata feature vector, and the corresponding metadata availability mask.
After Stage~A  training is complete, the selected Wavelet-Fusion VAE compressor is frozen. Real MRI volumes are encoded into latent tensors, the diffusion model is trained in this compressed space, and generated latent samples are mapped back to image space using the corresponding decoder.
This design follows standard latent diffusion practice while introducing the proposed Wavelet-Fusion VAE compressor and explicit conditioning of the latent diffusion model on target modality and metadata \parencite{rombach-LDM-2022,Pinaya-LDM3DBrain-2022,Guo-Maisi-2025}.

Figure~\ref{fig:wfdm-architecture} provides an overview of the two-stage WFDM pipeline.
Panel~A summarizes the proposed Wavelet-Fusion VAE compression stage, in which the input MRI volume is first decomposed using a two-stage wavelet decomposition.
The stage-1 projected wavelet features are provided as input to the early encoder stage.
A second wavelet decomposition is then applied to the selected subset of the stage-1 coefficients to obtain the stage-2 wavelet features.
After projection, these stage-2 features are fused into a deeper encoder stage.
On the decoder side, the reconstructed stage-2 wavelet coefficients are first inverted to recover the selected stage-1 subband representation. This recovered representation is then fused with the reconstructed stage-1 coefficients, and a final inverse wavelet transform reconstructs the output volume.
Panel~B summarizes latent diffusion in the learned Wavelet-Fusion VAE space.
During training, real MRI volumes are encoded into latent representations and noise is added progressively in the forward diffusion process. During inference, sampling starts from Gaussian noise and iteratively denoises in latent space before decoding back to MRI space.

\begin{figure}[p]
  \centering
  \includegraphics[
    width=0.98\textwidth,
    keepaspectratio
  ]{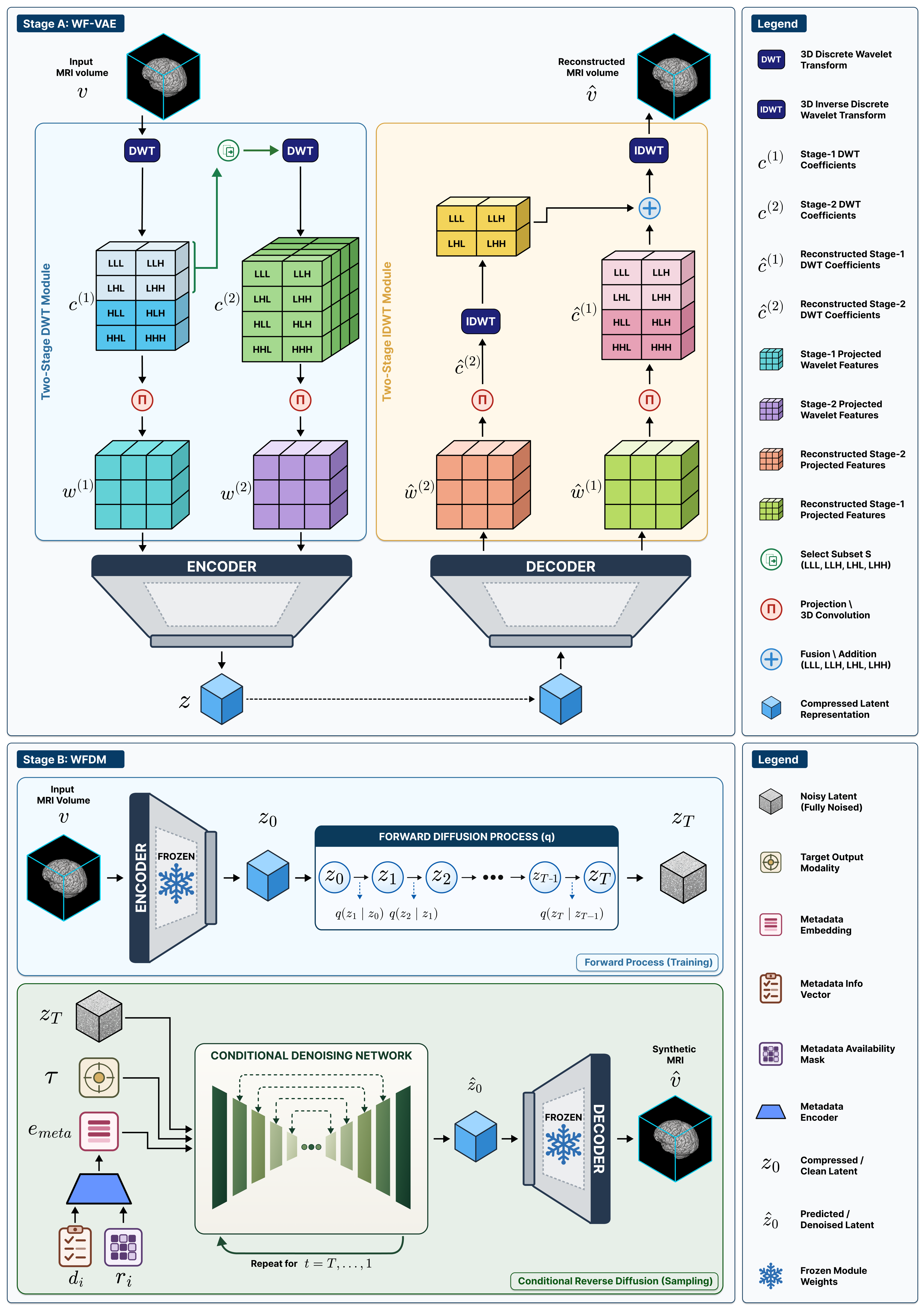}
\end{figure}

\clearpage

\begin{center}
\begin{minipage}{0.95\textwidth}
\captionof{figure}{
  Architecture of the proposed Wavelet-Fusion Diffusion Model (WFDM).
  \textbf{(A) Stage A: Wavelet-Fusion VAE (WF-VAE) compression.}
  A one-level 3D Haar discrete wavelet transform (DWT) is first applied to the input MRI volume, yielding eight subbands (see Equation~\ref{eq:wavelet-subbands}).
  These stage-1 coefficients are concatenated and then projected by a convolutional block to produce the stage-1 wavelet features $w^{(1)}$ (see Equation~\ref{eq:wavelet-stage1-features}), which are used as direct input to the encoder.
  A lower-frequency subset of the first-stage coefficients is transformed again by a second DWT stage, producing the stage-2 coefficients, which are projected to stage-2 wavelet features $w^{(2)}$ (see Equation~\ref{eq:wavelet-stage2-features}) and fused into a deeper encoder stage.
  The decoder reconstructs both stage-2 and stage-1 coefficient representations. The stage-2 representation is first inverted to recover the selected first-stage subset, which is fused with the reconstructed stage-1 coefficients before the final inverse DWT reconstructs the output MRI volume.
  \textbf{(B) Stage B: latent diffusion in the selected Wavelet-Fusion VAE space.}
  Real MRI volumes are encoded into latent tensors $z_0$, progressively noised during the forward diffusion process for training, and denoised by a conditional 3D U-Net during the reverse diffusion process.
  During inference, sampling begins from Gaussian noise $z_T \sim \mathcal{N}(0,I)$ and is denoised by the denoising network. The final denoised latent $\hat{z}_0$ is decoded by the selected Wavelet-Fusion VAE decoder to generate the synthetic MRI output.
}
\label{fig:wfdm-architecture}
\end{minipage}
\end{center}


\subsubsection{Stage A: Wavelet-Fusion VAE}
\label{sec:Methodology-WF-VAE}
Three-dimensional structural MRI contains fine anatomical structure, including tissue boundaries and small-scale details, that can be degraded by compression.
To better preserve anatomical detail within a compact latent representation, we augment the autoencoder with Wavelet-Fusion modules that introduce wavelet domain features at multiple stages into the autoencoder.
Let $v \triangleq x_i^{(m)}$, with $m \in \mathcal{M}_i^{\mathrm{avail}}$, denote a preprocessed single-modality input volume.
The encoder $E_{\phi}$ produces the parameters of a diagonal Gaussian approximate posterior over the latent representation:
\[
 q_{\phi}(z\mid v)=\mathcal{N}\big(\mu_{\phi}(v),\operatorname{diag}(\sigma_{\phi}^{2}(v))\big),
\]
where $z$ denotes the latent variable, $\mu_{\phi}(v)$ is the latent mean tensor, and $\sigma_{\phi}^{2}(v)$ is the latent variance tensor.
In our implementation, $\mu_{\phi}(v)$ and $\log \sigma_{\phi}^{2}(v)$ are predicted by separate $1\times1\times1$ convolutional layers.
The log-variance is clamped for numerical stability, and the corresponding standard deviation tensor is recovered as
\[
\sigma_{\phi}(v)=\exp\!\left(\tfrac{1}{2}\log \sigma_{\phi}^{2}(v)\right).
\]

A latent sample is then obtained using the reparameterization trick:
\[
z=\mu_{\phi}(v)+\sigma_{\phi}(v)\odot\varepsilon,
\qquad \varepsilon\sim\mathcal{N}(0,I).
\]

The decoder reconstructs the input as $\hat{v} = D_\psi(z)$. With a downsampling factor of $f=4$ and four latent channels, the latent tensor has shape
\[
 z \in \mathbb{R}^{4\times (H/4)\times(W/4)\times(D/4)}
 = \mathbb{R}^{4\times 48\times 56\times 40}
 \quad\text{for }(H,W,D)=(192,224,160).
\]

The autoencoder is trained with a weighted combination of voxel-space reconstruction, perceptual similarity, Kullback-Leibler divergence (KL) regularization, and an adversarial objective based on a patch discriminator, following the general VAE-GAN compression strategy used in previous latent diffusion pipelines for medical imaging \parencite{Pinaya-LDM3DBrain-2022,Guo-Maisi-2025}.
The voxel-space reconstruction loss term is
\[
 \mathcal{L}_{\mathrm{recon}}(v,\hat v) = \|v-\hat v\|_1.
\]

For the diagonal Gaussian posterior, the KL regularizer is
\[
 \mathcal{L}_{\mathrm{KL}}
 =
 D_{\mathrm{KL}}\!\big(q_\phi(z\mid v)\,\|\,\mathcal{N}(0,I)\big)
 =
 \frac{1}{2}\sum_{k}\Big(\mu_k^2 + \sigma_k^2 - \log(\sigma_k^2) - 1\Big),
\]
where the sum runs over all latent elements \parencite{kingma-VAE}.
To further improve perceptual fidelity, we also include a feature reconstruction loss based on a fixed pretrained feature extractor $\Phi(\cdot)$.
This encourages similarity in deep feature space beyond voxel-wise comparisons \parencite{Johnson-PrepLossST-2016}.
The perceptual feature reconstruction loss is
\[
 \mathcal{L}_{\mathrm{perc}}(v,\hat v)
 =
 \big\|\Phi(v)-\Phi(\hat v)\big\|_1.
\]

We further introduce a patch-based discriminator \(D_{\omega}\) during compression training.
For a real volume \(v\) and its reconstruction \(\hat v\), the discriminator loss is
\[
 \mathcal{L}_{\mathrm{adv}}^{D}(v,\hat v)
 =
 -\frac{1}{2}\Big(
 \mathbb{E}\big[\log D_\omega(v)\big]
 +
 \mathbb{E}\big[\log\big(1-D_\omega(\hat v)\big)\big]
 \Big),
\]
and the generator-side adversarial term is
\[
 \mathcal{L}_{\mathrm{adv}}^{G}(\hat v)
 =
 -\mathbb{E}\big[\log D_\omega(\hat v)\big].
\]
This adversarial term encourages perceptually sharper and more realistic reconstructions \parencite{goodfellow-GAN-2020,Guo-Maisi-2025}.
The total autoencoder and discriminator objectives are
\[
 \mathcal{L}_{VAE}
 =
 \mathcal{L}_{\mathrm{recon}}
 +\lambda_{\mathrm{KL}}\mathcal{L}_{\mathrm{KL}}
 +\lambda_{\mathrm{perc}}\mathcal{L}_{\mathrm{perc}}
 +\lambda_{\mathrm{adv}}\mathcal{L}_{\mathrm{adv}}^{G},
\]
and
\[
 \mathcal{L}_{D}
 =
 \mathcal{L}_{\mathrm{adv}}^{D}.
\]
Training alternates between updating the autoencoder parameters \((\phi,\psi)\) by minimizing \(\mathcal{L}_{VAE}\) with \(\omega\) fixed, and updating the discriminator parameters \(\omega\) by minimizing \(\mathcal{L}_{D}\) with \((\phi,\psi)\) fixed \parencite{Pinaya-LDM3DBrain-2022,Guo-Maisi-2025}.

\subsubsection{Wavelet-Fusion module}
To preserve fine anatomical detail under latent compression, we incorporate a two-stage Wavelet-Fusion feature pathway within the autoencoder.
The overall structure of the two-stage Wavelet-Fusion pathway is illustrated in Figure~\ref{fig:wfdm-architecture}.
At each stage, the pathway applies the 3D Haar discrete wavelet transform (DWT), which decomposes the input representation into low-frequency and high-frequency subbands along the three spatial dimensions.
Let $W(\cdot)$ and $W^{-1}(\cdot)$ denote the forward and inverse 3D Haar DWT and IDWT, respectively.
The transform applies separable low-pass and high-pass filters along the three spatial axes,
\[
\ell=\tfrac{1}{\sqrt{2}}[1,1],\qquad h=\tfrac{1}{\sqrt{2}}[-1,1],
\]
and decomposes the input volume into eight subbands at half spatial resolution.
This decomposition separates low-frequency structure from high-frequency detail and represents them as wavelet subbands at reduced spatial resolution \parencite{Friedrich-WDM-2024}.
A one-level DWT yields eight subbands:
\begin{equation}
W(v) = \big(v^{LLL},\,v^{LLH},\,v^{LHL},\,v^{LHH},\,v^{HLL},\,v^{HLH},\,v^{HHL},\,v^{HHH}\big),
\label{eq:wavelet-subbands}
\end{equation}
where each superscript records whether the low pass or high pass branch is selected along each of the three spatial axes.
In the first wavelet stage, the eight subbands are concatenated channel-wise:
\[
c^{(1)}=\operatorname{cat}\!\big(W(v)\big).
\]
The concatenated first-stage wavelet coefficients are mapped to a learned feature representation:
\begin{equation}
w^{(1)}=\Pi_{1}\!\left(c^{(1)}\right),
\label{eq:wavelet-stage1-features}
\end{equation}
which is used as the encoder input.
In the second wavelet stage, the selected subset of the stage-1 coefficients is transformed again:
\[
 c^{(2)}=\operatorname{cat}\!\Big(W\big(S(c^{(1)})\big)\Big),
\]
where $S(\cdot)$ denotes the subset selection operator that selects the lower frequency channels from the stage-1 stacked coefficients according to the subband ordering in Equation~\ref{eq:wavelet-subbands}.
These coefficients are then projected to
\begin{equation}
w^{(2)}=\Pi_{2}\!\left(c^{(2)}\right),
\label{eq:wavelet-stage2-features}
\end{equation}
which is fused with the corresponding second-stage encoder features.
On the decoder side, the network predicts stage-2 ($\hat c^{(2)}$) and stage-1 ($\hat c^{(1)}$) coefficient stacks.
The stage-2 coefficients are first inverted to recover the selected low-frequency subset of the stage-1 representation.
The recovered coefficients are then fused with the corresponding stage-1 predicted coefficients to obtain the reconstructed stage-1 coefficient stack $\tilde{c}^{(1)}$, and the final image is obtained by applying the stage-1 inverse transform:
\[
\hat{v}=W^{-1}\!\left(\tilde{c}^{(1)}\right).
\]
This design introduces wavelet features into the autoencoder, complementing the learned convolutional representation with information from both low-frequency structure and high-frequency detail.

\subsubsection{Latent scaling}
To improve the stability of diffusion training, we rescale the autoencoder latents using a single global factor estimated from the training set so that the latent tensors processed by the diffusion model have approximately unit variance.
Let $\tilde{z}$ denote the raw autoencoder latent and let $s_z$ denote the latent scaling factor estimated from the training split.
The scaled latent used by the diffusion model is
\[
z = s_z \tilde{z}.
\]
Throughout the remainder of this section, $z$ denotes this scaled latent representation.
The corresponding inverse scaling, $\tilde{z}=z/s_z$, is applied after the diffusion denoising process, immediately before decoding back to image space.

\subsubsection{Stage B: Latent diffusion in Wavelet-Fusion VAE space}
The latent diffusion stage is summarized schematically in Figure~\ref{fig:wfdm-architecture}.
In Stage~B, we train a conditional denoising diffusion model in the latent space learned by the selected Wavelet-Fusion VAE compressor.
Let $z_0$ denote the scaled clean latent representation of an input volume.
The forward Markov process progressively adds Gaussian noise to $z_0$ over $T$ diffusion steps, while a denoising network $\epsilon_{\theta}$ is trained to predict the injected noise at arbitrary timesteps.
During inference, sampling begins from Gaussian noise $z_T\sim\mathcal{N}(0,I)$ and proceeds by iterative denoising to obtain a clean scaled latent sample. 
The latent scaling is then reversed, and the resulting latent is mapped back to image space by the corresponding decoder \parencite{ho-DDPM-2020}. 
The forward noising process has a predefined variance schedule $\{\beta_t\}_{t=1}^{T}$ with
\[
 \alpha_t = 1-\beta_t,\qquad \bar{\alpha}_t=\prod_{j=1}^{t}\alpha_j, \qquad \bar{\alpha}_0=1.
\]
The Markov forward transition is
\[
 q(z_t\mid z_{t-1})
 =\mathcal{N}\!\big(\sqrt{\alpha_t}\,z_{t-1},\ (1-\alpha_t)I\big),
\]
and the corresponding marginal distribution used to sample noisy latents during training is
\[
 q(z_t\mid z_0)=\mathcal{N}\!\big(\sqrt{\bar{\alpha}_t}\,z_0,\ (1-\bar{\alpha}_t)I\big),
 \qquad
 z_t=\sqrt{\bar{\alpha}_t}\,z_0+\sqrt{1-\bar{\alpha}_t}\,\epsilon,\quad \epsilon\sim\mathcal{N}(0,I).
\]
The denoising network $\epsilon_{\theta}$ predicts the injected noise conditioned on $c_{i,\tau}$:
\[
 \hat{\epsilon}=\epsilon_{\theta}(z_t,t,c_{i,\tau}),
\]
and is trained with the standard objective for predicting noise
\[
 \mathcal{L}_{\mathrm{diff}}=\|\epsilon-\hat{\epsilon}\|_2^2.
\]

At inference time, sampling starts from $z_T\sim\mathcal{N}(0,I)$ and follows the reverse process
\[
 p_{\theta}(z_{t-1}\mid z_t,c_{i,\tau})=\mathcal{N}\!\big(\mu_{\theta}(z_t,t,c_{i,\tau}),\ \tilde{\beta}_t I\big),
 \qquad
 \tilde{\beta}_t=\frac{1-\bar{\alpha}_{t-1}}{1-\bar{\alpha}_t}\,\beta_t,
\]
with mean parameterized by the predicted noise:
\[
 \mu_{\theta}(z_t,t,c_{i,\tau})=
 \frac{1}{\sqrt{\alpha_t}}
 \left(
 z_t-\frac{\beta_t}{\sqrt{1-\bar{\alpha}_t}}\,\epsilon_{\theta}(z_t,t,c_{i,\tau})
 \right).
\]
Iterating from $t=T$ to $1$ yields a generated scaled clean latent estimate $\hat{z}_0$. 
Before decoding, inverse latent scaling is applied, and the generated MRI volume is reconstructed as
\[
\hat{v}=D_{\psi}\!\left(\frac{\hat{z}_0}{s_z}\right).
\]

\subsubsection{Modality and metadata conditioning}
\label{sec:Methodology-Modality-Metadata-Cond}

The latent diffusion denoiser is conditioned jointly on the target MRI modality and the associated demographic and clinical metadata.

\textbf{Target-modality conditioning:}
The target modality $\tau$ is treated as a discrete class label indicating the desired output contrast.
A learnable modality embedding is computed from this label and added to the timestep embedding.
This design encourages the network to learn a shared anatomical prior while conditioning generation on the target modality label to control the appearance of the requested MRI contrast.

\textbf{Metadata conditioning and metadata availability encoding:}
For each subject entry $i$, the dataset transform converts the raw metadata dictionary into a metadata representation with a fixed schema, consisting of a values vector $d_i$ and a corresponding metadata availability mask $r_i$.
Continuous metadata fields are normalized as scalar values, while categorical fields and clinical yes/no variables are converted to integer identifiers.
The mask $r_i$ records which metadata fields are observed, allowing the metadata encoder to distinguish unavailable fields from valid observed values, including explicit negative responses.

A dedicated metadata encoder maps $(d_i,r_i)$ to a metadata embedding of fixed width $e_{\mathrm{meta}}$.
In this encoder, normalized numeric variables are used directly, categorical and clinical variables are represented through learned embeddings, and the metadata availability mask is incorporated to distinguish observed covariates from unavailable fields.
The resulting metadata embedding $e_{\mathrm{meta}}$ is projected to the U-Net embedding width and concatenated with the combined timestep and modality embedding.
The combined vector is then injected into the 3D U-Net.
This conditioning pathway incorporates both the metadata feature vector and the metadata availability mask, allowing the model to use available demographic and clinical context while conditioning synthesis on the target modality \parencite{Yasinzai-BrainScape-2025}.
\subsubsection{Training protocol}

Training is performed in two stages.
In Stage~A, the autoencoder is trained on standardized 3D MRI volumes using the reconstruction, KL, perceptual, and adversarial objectives defined in Section~\ref{sec:Methodology-WF-VAE}.
Optimization alternates between autoencoder and discriminator updates, together with learning-rate scheduling to stabilize convergence.

In Stage~B, the trained autoencoder is frozen, and the latent diffusion model is trained on the corresponding scaled latent representations.
The denoiser is optimized using the standard $\ell_2$ noise-prediction objective, while conditioning is composed of modality and metadata as described in Section~\ref{sec:Methodology-Modality-Metadata-Cond}.
To mitigate bias toward the most abundant target modalities, modality imbalance is addressed through weighted sampling, ensuring rarer target modalities are adequately represented during training.

\subsubsection{Evaluation methodology}

We use two distinct evaluation settings corresponding to the two stages of the framework.
First, we evaluate the autoencoder as a compression model by measuring how faithfully it reconstructs input MRI volumes.
Second, we evaluate the diffusion model as a generator by assessing the distributional realism of its synthesized samples and their diversity.

\textbf{Autoencoder reconstruction fidelity:}
The autoencoder is evaluated using paired reconstruction metrics that compare each real volume $v$ with its reconstruction $\hat{v}$.
We report peak signal-to-noise ratio (PSNR), structural similarity index measure (SSIM), multi-scale structural similarity index measure (MS-SSIM), learned perceptual image patch similarity (LPIPS), and mean squared error (MSE) in the preprocessed image space to quantify compression reconstruction fidelity.

\textbf{Diffusion sample realism:}
To evaluate the distributional realism of the generated images, we use the \textit{medmetric} Python package, which standardizes feature extraction and feature-space evaluation for 3D medical images using a fixed pretrained MedicalNet backbone \parencite{Yasinzai-MedMetricGitHub}.
In our experiments, real and synthetic volumes are z-score normalized before feature extraction, consistent with the intended MedicalNet workflow for computing FID and MMD metrics.

Let $F$ denote the fixed MedicalNet feature extractor, which extracts features in MedicalNet feature space rather than ImageNet Inception feature space.
For a 3D MRI volume $v$, we define its pooled MedicalNet feature representation as
\[
f = F(v) \in \mathbb{R}^{d}.
\]

FID and MMD are then computed using these MedicalNet feature representations rather than directly on voxel intensities.
Given generated feature vectors $G=\{g_a\}_{a=1}^{n_g}$ and real feature vectors $R=\{r_b\}_{b=1}^{n_r}$, let $(\mu_g,\Sigma_g)$ and $(\mu_r,\Sigma_r)$ denote their empirical means and covariance matrices, respectively.
The Fr\'echet distance is defined as
\[
\mathrm{FID}=
\|\mu_r-\mu_g\|_2^2+
\operatorname{Tr}\!\Big(\Sigma_r+\Sigma_g-2(\Sigma_r\Sigma_g)^{1/2}\Big).
\]

We additionally compute maximum mean discrepancy (MMD) using the same MedicalNet feature vectors.
The unbiased estimator is

\[
\widehat{\mathrm{MMD}}^{2}(G,R)=
\frac{1}{n_g(n_g-1)}\!\sum_{a\neq b}\!k(g_a,g_b)
+\frac{1}{n_r(n_r-1)}\!\sum_{a\neq b}\!k(r_a,r_b)
-\frac{2}{n_g n_r}\!\sum_{a=1}^{n_g}\sum_{b=1}^{n_r}\!k(g_a,r_b).
\]

The kernel is a Gaussian mixture:
\[
 k(x,y)=\sum_{\ell=1}^{L} w_\ell 
 \exp\!\left(-\frac{\|x-y\|_2^2}{2\sigma_\ell^2}\right),
 \qquad 
 w_\ell \ge 0,\quad \sum_{\ell=1}^{L} w_\ell=1.
\]
Finally, the reported MMD value is computed as
\[
\mathrm{MMD}=\sqrt{\max(\widehat{\mathrm{MMD}}^{2},0)}.
\]
Lower FID and MMD indicate closer agreement between the generated and real feature distributions.

\textbf{Diffusion sample diversity:}
To assess diversity and detect mode collapse, we compute MS-SSIM between generated samples directly on generated volumes in image space.
For each modality or conditioning bin, we sample $K$ random pairs from $N$ generated volumes and compute the average similarity between generated samples:
\[
\mathrm{MS\text{-}SSIM}_{\mathrm{ff}}
=
\frac{1}{K}\sum_{k=1}^{K}
\mathrm{MS\text{-}SSIM}\!\left(\hat v^{(a_k)},\,\hat v^{(b_k)}\right),
\qquad a_k \neq b_k.
\]
Lower values indicate lower self-similarity among generated samples and therefore greater sample diversity.
All synthesis metrics are computed using a fixed evaluation protocol on the BrainScape test split, with generated samples conditioned on target modalities and metadata drawn from the test split.
Results are reported by modality, together with pooled summaries \parencite{Yasinzai-BrainScape-2025}.

\FloatBarrier
\section{Experiments}
\label{sec:experiments}

\subsection{Experiment A: Autoencoder benchmarking}

The first experiment evaluated candidate latent compressors for 3D MRI synthesis.
The goal was to identify a representation that preserved fine anatomical detail under compression while remaining practical for large-scale 3D synthesis.
We compared latent compressors used in prior medical-image generation pipelines, including ALDM VQ-GAN \parencite{Kim-ALDM-2024}, BrainSynth VQ-VAE \parencite{Peng-BrainSynth-2024}, the LDM-VAE \parencite{Pinaya-LDM3DBrain-2022}, and the MAISI VAE-GAN \parencite{Guo-Maisi-2025}, all of which were externally pretrained on datasets other than BrainScape.

We also evaluated autoencoders trained on BrainScape, including a VAE-GAN baseline, Wavelet VAE, and the two proposed Wavelet-Fusion variants: Wavelet-Fusion VAE (WF-VAE) and Wavelet-Fusion Self-Attention VAE (WF-SA-VAE). 
The Wavelet VAE uses a fixed two-stage wavelet decomposition to convert each input MRI volume into a 64-channel stage-2 wavelet representation before latent encoding.
This 64-channel wavelet-domain representation is then passed to a simplified autoencoder bottleneck that compresses it into a 4-channel latent representation.
This design is intended to reduce the burden on the autoencoder to learn spatial downsampling entirely from voxel space, because the Wavelet VAE input already contains structured low-frequency and high-frequency information.
This model evaluates whether a wavelet-domain input with a lightweight learned autoencoder bottleneck is sufficient for latent MRI reconstruction. 
In contrast, our proposed WF-VAE and WF-SA-VAE models retain the learnable autoencoder downsampling pathway and fuse first-stage and second-stage wavelet features into the encoder, as described in Section~\ref{sec:Methodology-WF-VAE}.
WF-SA-VAE differs from WF-VAE by including self-attention blocks in the VAE's deepest bottleneck stage.

In addition to the learned compressors, we computed the reconstruction performance of one-stage wavelet transforms used in WDM \parencite{Friedrich-WDM-2024} and two-stage wavelet transforms as nearly lossless reconstruction references, both of which do not require training. 
All models were evaluated by reconstructing held-out 3D BrainScape volumes.
Reconstruction quality was assessed both for each modality and for the pooled results using LPIPS, SSIM, MS-SSIM, PSNR, and MSE.
This experiment was designed to evaluate the performance of our proposed Wavelet-Fusion latent compressors compared with both BrainScape-trained autoencoder baselines and externally pretrained compressor references.
The best performing learned compressor trained on BrainScape was then selected as the latent representation for the proposed diffusion model in Experiments~B, C, and D.

\subsection{Experiment B: Diffusion model benchmarking}

The second experiment evaluated candidate generative models for multimodal 3D MRI synthesis on BrainScape. 
Our proposed model, WFDM, performs diffusion in the latent space of WF-SA-VAE and is conditioned on the target output modality together with a metadata embedding derived from demographic and clinical metadata and the corresponding metadata availability mask.
The primary internal baseline follows the same overall latent diffusion formulation and uses the same modality and metadata conditioning interface, but replaces the Wavelet-Fusion latent backbone with MAISI VAE-GAN.

We also report results from pretrained external generators for modalities where direct evaluation is possible, including LDM \parencite{Pinaya-LDM3DBrain-2022}, WDM \parencite{Friedrich-WDM-2024}, HA-GAN \parencite{Sun-HAGAN-2022}, and Med-DDPM \parencite{Dorjsembe-MedDDPM-2024}. 
Table~\ref{tab:baseline-summary} summarizes the representative generators, their conditioning strategies, synthesis domains, sampling configurations, training sources, and modality coverage.
Evaluation of the benchmarked models was performed using the \textit{medmetric} package \parencite{Yasinzai-MedMetricGitHub}. 
For distributional realism, FID and MMD were computed in the shared MedicalNet feature space. 
For diversity, MS-SSIM was computed between generated volumes in image space as a diversity proxy. 
Lower FID and MMD indicate closer alignment between real and synthetic feature distributions, whereas lower MS-SSIM between generated samples indicates lower self-similarity and therefore greater sample diversity.

\begin{table*}
\centering
\scriptsize
\begin{threeparttable}
\caption{Summary of evaluated synthesis models.\tnote{$a$}}
\label{tab:baseline-summary}
\setlength{\tabcolsep}{3pt}
\begin{tabular}{@{}lllllll@{}}
\toprule
Model       & Conditioning         & Synthesis domain       & Scheduler    & Steps  & Training source         & Output modalities \\
\midrule
WFDM        & Modality + metadata  & Latent (WF-SA-VAE)     & RFlow        & 50     & BrainScape              & T1w, T2w, T1Gd, FLAIR \\
MAISI       & Modality + metadata  & Latent (MAISI VAE-GAN) & RFlow        & 50     & BrainScape              & T1w, T2w, T1Gd, FLAIR \\
LDM         & None (unconditional) & Latent (LDM-VAE)       & DDIM         & 50     & UK Biobank (pretrained) & T1w \\
Med-DDPM    & Segmentation mask    & Pixel                  & DDPM         & 250    & BraTS2021 (pretrained)  & T1w, T2w, T1Gd, FLAIR \\
WDM         & None (unconditional) & DWT coefficient space  & DDPM         & 1000   & BraTS2023 (pretrained)  & T1w \\
HA-GAN      & None (unconditional) & Pixel                  & N/A (GAN)    & N/A    & GSP (pretrained)        & T1w \\
\bottomrule 
\end{tabular}

\begin{tablenotes}[flushleft]\footnotesize
\item[${a}$]
Columns summarize each model's conditioning input, synthesis domain, scheduler, sampling steps, training source, and output modalities evaluated in this study. 
WFDM and MAISI were trained on BrainScape in this study, whereas LDM, Med-DDPM, WDM, and HA-GAN are externally pretrained reference models. 
\end{tablenotes}
\end{threeparttable}
\end{table*}

\subsection{Experiment C: Sampling and loss ablation}
\label{sec:Experiment-sampling-ablation}

In the third experiment, we performed an ablation study on the proposed WFDM model family to evaluate how the generative formulation, noise schedule, prediction objective, and loss norm affect synthesis quality. 
Specifically, we compared DDPM-based sampling against Rectified Flow (RFlow) as alternative generative formulations. 
Within the DDPM setting, we further ablated the noise schedule by comparing cosine and scaled-linear $\beta$ schedules. 
For the prediction objective, DDPM variants used the $v$-prediction parameterization, while the RFlow variant used the corresponding velocity/flow prediction objective. 
Across selected configurations, we compared $\ell_1$ and $\ell_2$ regression losses to evaluate their impact.
This experiment compared DDPM and RFlow sampling configurations to quantify how sampler choice affects synthesis quality and computational cost in latent diffusion.
The default configuration for the final proposed model was selected by prioritizing lower pooled FID and MMD while verifying that MS-SSIM between generated samples remained competitive and did not indicate diversity collapse.

\subsection{Experiment D: Inference-time efficiency benchmark}
\label{sec:Experiment-inference-efficiency}

To assess the practical deployment cost of the evaluated generative models, we performed an inference-time efficiency benchmark on a single NVIDIA H100 GPU on New Zealand eScience Infrastructure (NeSI). 
All benchmarks were run on a single HPC node with 8 CPU cores and 64\,GB host memory using CUDA~12.6.3, cuDNN~9.5.1.17, Python~3.11.13, and PyTorch~2.4.1. 
To isolate model execution performance, inference was performed with a batch size of 1, and the reported timing intervals measure only the model inference time, excluding disk I/O and data loading overhead.

For each benchmarked model, we first ran three warm-up generations to reduce initialization overhead and then recorded 30 timed generations.
Timing was measured using synchronized GPU wall-clock timing, immediately before and after each inference call. 
For each configuration, we report the mean, median, and standard deviation of the time required to generate a single 3D MRI volume, together with the peak GPU memory usage and output tensor size.

This experiment is reported in two parts. 
The main efficiency benchmark compares all evaluated generators under the same measurement protocol. 
We also report a second efficiency table for the WFDM sampling and loss ablation study in Section~\ref{sec:Experiment-sampling-ablation}, which quantifies how the generative formulation, noise schedule, prediction objective, loss norm, and associated number of inference steps affect practical runtime within the same WFDM model family.

\FloatBarrier
\section{Results}
\label{sec:results}
\subsection{Compressor reconstruction results}
Among the BrainScape-trained learned compressors, WF-SA-VAE achieved the strongest overall performance, with the lowest LPIPS (0.0634) and MSE (0.000885), and the highest SSIM (0.9646), MS-SSIM (0.9947), and PSNR (30.79). 
Across all learned compressors, MAISI VAE-GAN remained the strongest external reference, achieving LPIPS of 0.0415, SSIM of 0.9675, MS-SSIM of 0.9953, PSNR of 32.55, and MSE of 0.000621 (see Table~\ref{tab:ae-pooled} for full results).
Although the externally pretrained MAISI VAE-GAN achieved the best absolute reconstruction metrics overall, WF-SA-VAE was the strongest learned compressor trained on BrainScape and substantially outperformed the matched VAE-GAN baseline across all pooled metrics.
WF-SA-VAE was therefore selected as the proposed latent compressor for Experiments~B, C, and D, and the subsequent diffusion experiment showed that the WF-SA-VAE latent space supported the strongest overall synthesis performance.

\begin{table*}[t]
\centering
\scriptsize
\begin{threeparttable}
\caption{Autoencoder reconstruction metrics pooled across modalities.\tnote{$a$}}
\label{tab:ae-pooled}
\setlength{\tabcolsep}{2.5pt}
\renewcommand{\arraystretch}{1.1}
\begin{tabular}{@{}llccccccc@{}}
\toprule
Model & Training dataset & Comp & Chans & LPIPS $\downarrow$ & SSIM $\uparrow$ & MS-SSIM $\uparrow$ & PSNR $\uparrow$ & MSE $\downarrow$ \\
\midrule
ALDM VQ-GAN \parencite{Kim-ALDM-2024} & BraTS2021 / IXI & --- & --- & 0.4739 & 0.7962 & 0.8243 & 16.86 & 0.506754 \\
BrainSynth VQ-VAE \parencite{Peng-BrainSynth-2024} & Multi-study T1w MRI & --- & --- & 0.6270 & 0.6631 & 0.8106 & 19.91 & 0.010698 \\
LDM-VAE \parencite{Pinaya-LDM3DBrain-2022} & UK Biobank T1w MRI & 1/8 & 3 & 0.2334 & 0.8641 & 0.9544 & 22.98 & 0.006548 \\
VAE-GAN (BrainScape) & BrainScape & 1/4 & 4 & 0.1418 & 0.6121 & 0.9828 & 27.32 & 0.002556 \\
MAISI VAE-GAN \parencite{Guo-Maisi-2025} & Public CT+MRI & 1/4 & 4 & 0.0415 & 0.9675 & 0.9953 & 32.55 & 0.000621 \\
Wavelet VAE & BrainScape & 1/4 & 4 & 0.2917 & 0.6273 & 0.9681 & 24.46 & 0.003785 \\
WF-VAE & BrainScape & 1/4 & 4 & 0.0798 & 0.9580 & 0.9927 & 29.65 & 0.001182 \\
WF-SA-VAE & BrainScape & 1/4 & 4 & \textbf{0.0634} & \textbf{0.9646} & \textbf{0.9947} & \textbf{30.79} & \textbf{0.000885} \\
WDM-style DWT (1-stage) \parencite{Friedrich-WDM-2024} & N/A & 1/2 & 8 & 0.0000 & 1.0000 & 1.0000 & 82.63 & 0.000000 \\
DWT reference (2-stage) & N/A & 1/4 & 64 & 0.0000 & 1.0000 & 1.0000 & 76.61 & 0.000000 \\
\bottomrule
\end{tabular}
\begin{tablenotes}[flushleft]\footnotesize
\item[${a}$]
The \textit{Training dataset} column reports the source data used for each learned compressor. 
N/A denotes fixed, non-learnable representations based on wavelet transforms. 
\textit{Comp} column reports the spatial compression factor per dimension, for example, 1/4 means each spatial dimension is reduced by a factor of four. 
\textit{Chans} column reports continuous latent feature channels where applicable, and wavelet coefficient channels for latent representations based on fixed wavelet transforms. 
For VQ-based compressors, latent channel values are omitted because discrete codebook representations are not directly comparable with continuous VAE latent channels.
LPIPS, SSIM, MS-SSIM, PSNR, and MSE compare input and reconstructed MRI volumes. Lower is better for LPIPS/MSE and higher is better for SSIM/MS-SSIM/PSNR. 
Bold indicates the best learned compressor trained on BrainScape, excluding externally pretrained compressors and fixed wavelet references.
\end{tablenotes}
\end{threeparttable}
\end{table*}

\begin{figure*}[t]
  \centering
  \includegraphics[width=\textwidth]{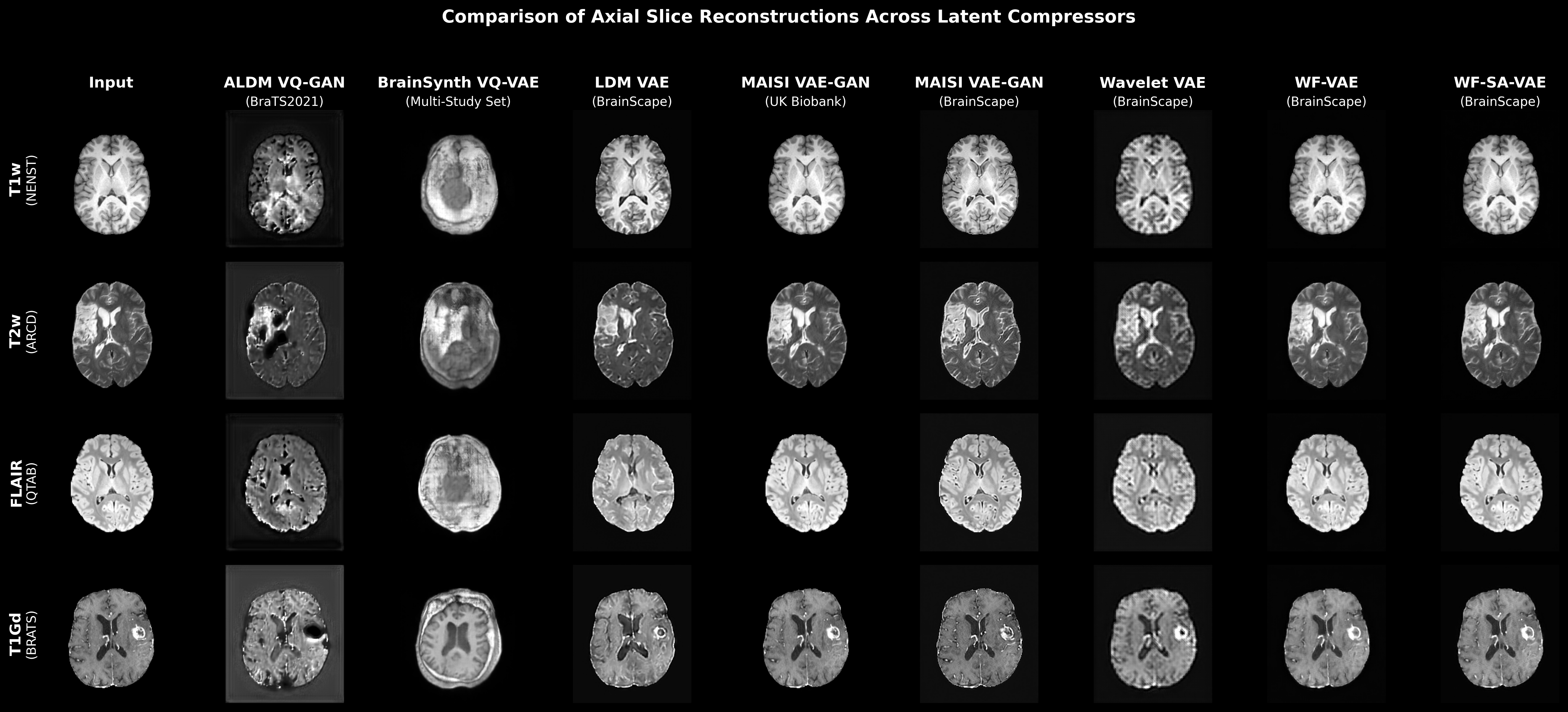}
  \caption{
  Qualitative comparison of axial-slice reconstructions across different latent compressors. 
  Each row shows a different MRI modality drawn from a different BrainScape source dataset: 
  FLAIR from QTAB (female, age 12 years), T1Gd from BraTS (tumor case), T1w from NENST (demographics unavailable), 
  and T2w from ARCD (male, age 56 years, with a recorded stroke label) \parencite{Yasinzai-BrainScape-2025}.
  The first column shows the preprocessed input image and the remaining columns show reconstructions from the pretrained external compressors 
  ALDM VQ-GAN \parencite{Kim-ALDM-2024}, BrainSynth VQ-VAE \parencite{Peng-BrainSynth-2024}, LDM-VAE \parencite{Pinaya-LDM3DBrain-2022}, and MAISI VAE-GAN \parencite{Guo-Maisi-2025}, 
  together with the BrainScape-trained VAE-GAN, Wavelet VAE, and the two proposed Wavelet-Fusion variants, WF-VAE and WF-SA-VAE.
  }
  \label{fig:ae-qualitative}
\end{figure*}

Figure~\ref{fig:ae-qualitative} provides a qualitative comparison of axial-slice reconstructions across the evaluated latent compressors. 
The figure includes four BrainScape test examples spanning different modalities, source datasets, demographic contexts, and available clinical labels, allowing reconstruction quality to be assessed under realistic cross-cohort heterogeneity \parencite{Yasinzai-BrainScape-2025}. 
The external pretrained ALDM VQ-GAN \parencite{Kim-ALDM-2024} and BrainSynth VQ-VAE \parencite{Peng-BrainSynth-2024} show visibly weaker anatomical fidelity and poorer reconstruction performance on these BrainScape examples. 
LDM-VAE \parencite{Pinaya-LDM3DBrain-2022} produces blurrier reconstructions and shows poorer recovery of fine anatomical detail.
The externally pretrained MAISI VAE-GAN \parencite{Guo-Maisi-2025} produces sharp and anatomically plausible reconstructions, consistent with its strong quantitative performance in Table~\ref{tab:ae-pooled}. 
Among the compressors trained on BrainScape, Wavelet VAE reconstructions are visibly weaker, with blurrier structures and poorer recovery of fine anatomical detail. 
The proposed Wavelet-Fusion variants preserve cortical boundaries, ventricular anatomy, and modality-specific contrast more accurately, with WF-SA-VAE producing the highest contrast and most anatomically coherent reconstructions among the learned compressors trained on BrainScape.
The T1Gd example remains the most challenging case. 
Although the Wavelet-Fusion variants provide reasonable reconstructions, the externally pretrained MAISI VAE-GAN produces a cleaner and sharper T1Gd reconstruction. 
This is consistent with the modality-specific results showing that T1Gd is the weakest modality for our proposed latent compressor.

The full modality-specific reconstruction tables are provided in Supplementary Material Appendix~A. 
Across T1w, T2w, T1Gd, and FLAIR, the results for each modality were consistent with the pooled comparison.
In every modality, WF-SA-VAE remained the strongest learned compressor trained on BrainScape, while WF-VAE was the next strongest variant and the simpler Wavelet VAE lagged behind the fusion-based designs. 
The external MAISI VAE-GAN reference remained strongest overall, although the gap to WF-SA-VAE was comparatively small in T1w, T2w, and FLAIR and became more pronounced in T1Gd. 
T1Gd was the most challenging reconstruction setting overall, as reflected by lower SSIM and higher LPIPS and MSE across nearly all learned models.

\subsection{Diffusion model synthesis results}

\begin{figure*}[t]
  \centering
  \includegraphics[width=\textwidth]{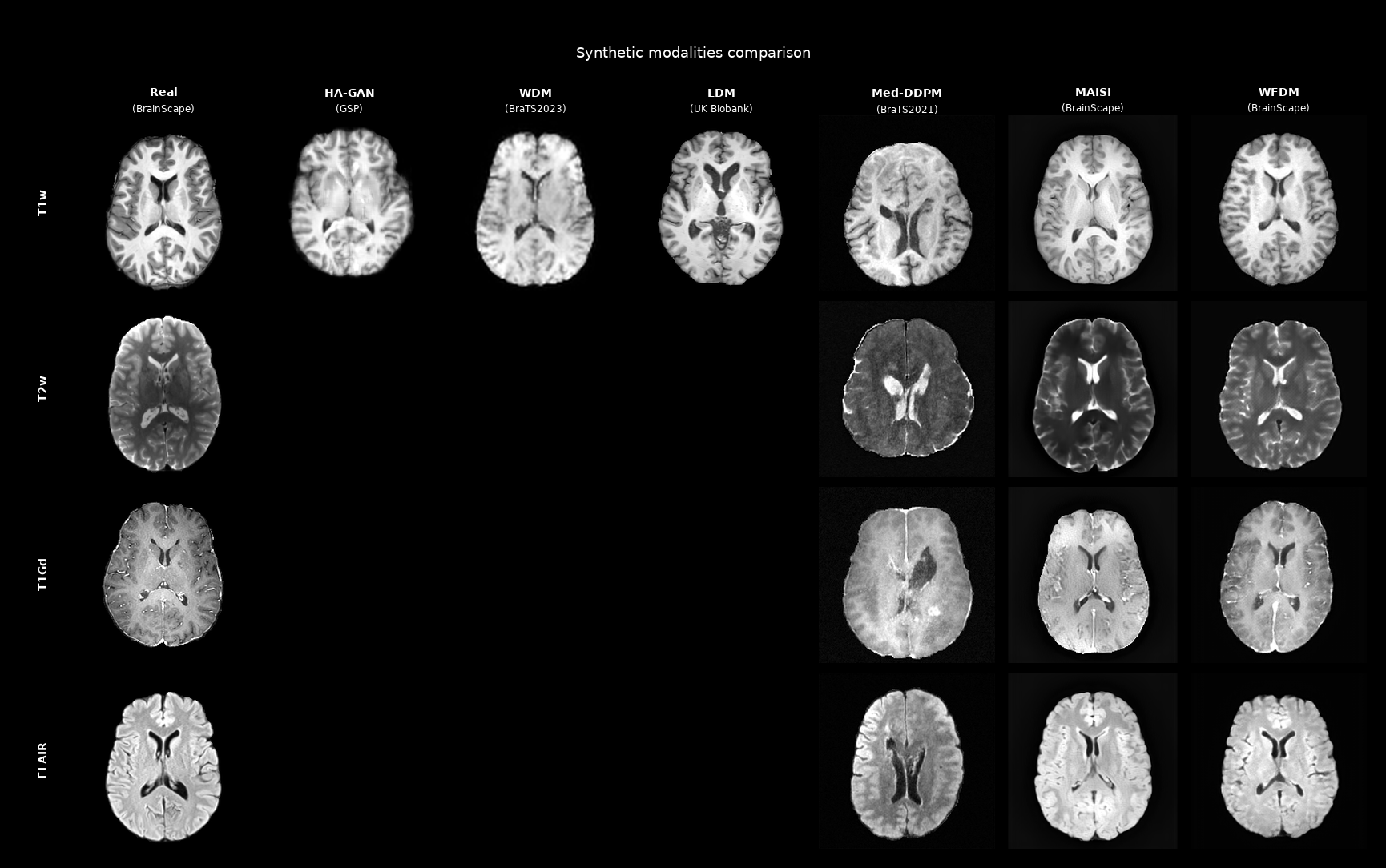}
  \caption{
  Qualitative axial-slice comparison across the four target modalities. 
  Rows correspond to T1w, T2w, T1Gd, and FLAIR MRI modalities, and columns show samples from each generator. 
  The synthetic MRI generators include HA-GAN \parencite{Sun-HAGAN-2022}, WDM \parencite{Friedrich-WDM-2024}, LDM \parencite{Pinaya-LDM3DBrain-2022}, Med-DDPM \parencite{Dorjsembe-MedDDPM-2024}, the BrainScape-trained MAISI internal baseline, and the proposed WFDM. 
  Empty entries indicate that the corresponding pretrained model was not configured to generate that modality.
  }
  \label{fig:diff-qualitative-axial}
\end{figure*}

Figure~\ref{fig:diff-qualitative-axial} provides a qualitative axial-slice comparison across all four target modalities. 
The figure also highlights the differences in modality coverage across the evaluated synthetic MRI generators.
Among the pretrained external baselines, HA-GAN \parencite{Sun-HAGAN-2022}, WDM \parencite{Friedrich-WDM-2024}, and LDM \parencite{Pinaya-LDM3DBrain-2022} are configured for T1w synthesis only, which is reflected by the missing T2w, T1Gd, and FLAIR entries in Figure~\ref{fig:diff-qualitative-axial}. 
By contrast, Med-DDPM \parencite{Dorjsembe-MedDDPM-2024}, the BrainScape-trained MAISI baseline, and the proposed WFDM generate all four target modalities.
This difference is not trivial as our goal is to develop a controllable multimodal MRI generator that remains effective in a heterogeneous pooled neuroimaging setting, rather than a model restricted to a single contrast or a narrowly defined cohort.

Qualitatively, the two latent diffusion models trained on BrainScape, MAISI and WFDM, produce the most consistent anatomy and contrast appearance across the four modalities.
Among the T1w-only external baselines, LDM is visually strongest, whereas HA-GAN and WDM are blurrier and recover less cortical detail. 
Although Med-DDPM is capable of generating all four modalities, its samples more often exhibit texture or striping artifacts, reduced sharpness, and weaker structural fidelity than the models trained on BrainScape. 
Overall, WFDM is visually strongest for T1w, T2w, and FLAIR, where it preserves ventricular morphology and contrast transitions more cleanly than the other baselines. 
T1Gd remains the hardest modality for WFDM, and the MAISI baseline generally produces a cleaner and sharper T1Gd sample, which is consistent with the quantitative modality-specific results in Appendix~B showing that the MAISI baseline outperforms the proposed model for this contrast.

WFDM achieved the best overall agreement with the real BrainScape distribution, with FID $=0.00404259$ and MMD $=0.147064$, improving over the matched internal MAISI baseline \parencite{Guo-Maisi-2025} (FID $=0.00549863$, MMD $=0.191766$). 
This corresponds to approximately $26\%$ lower pooled FID and $23\%$ lower pooled MMD, while diversity measured using MS-SSIM remained very similar between the two models. 
The proposed WFDM therefore generated 3D MRI samples that were better aligned with the real BrainScape distribution while maintaining comparable sample diversity.
Table~\ref{tab:diff-pooled} summarizes the pooled diffusion results. 
For the BrainScape-trained models and for Med-DDPM, metrics are pooled across all four generated modalities, whereas metrics for T1w-only external references are computed only on generated T1w samples.

The pretrained external baselines showed weaker evaluation metrics under this BrainScape evaluation protocol. 
LDM \parencite{Pinaya-LDM3DBrain-2022} and WDM \parencite{Friedrich-WDM-2024} both yielded larger FID and MMD values than the models trained on BrainScape, whereas HA-GAN \parencite{Sun-HAGAN-2022} remained far from the real BrainScape distribution despite showing higher diversity, as indicated by relatively low MS-SSIM between generated samples. 
These results show that diversity alone is not sufficient and must be considered together with distributional realism.

\begin{table}[t]
\centering
\begin{threeparttable}
\caption{Diffusion synthesis metrics for each evaluated generator.\tnote{$a$}}
\label{tab:diff-pooled}
\setlength{\tabcolsep}{10pt}%
\renewcommand{\arraystretch}{1.12}
\begin{tabular}{@{}llccc@{}}
\toprule
Model     & Output modalities      & FID $\downarrow$ & MMD $\downarrow$ & MS-SSIM $\downarrow$ \\
\midrule
WFDM      & T1w, T2w, T1Gd, FLAIR  & 0.00404259       & 0.147064         & 0.754384 \\
MAISI     & T1w, T2w, T1Gd, FLAIR  & 0.00549863       & 0.191766         & 0.757653 \\
LDM       & T1w                    & 0.01900610       & 0.411644         & 0.825484 \\
Med-DDPM  & T1w, T2w, T1Gd, FLAIR  & 0.05288810       & 0.341749         & 0.593099 \\
WDM       & T1w                    & 0.05860420       & 0.444065         & 0.733082 \\
HA-GAN    & T1w                    & 1.85993000       & 0.934294         & 0.694228 \\
\bottomrule
\end{tabular}
\begin{tablenotes}[flushleft]\footnotesize
\item[${a}$]
For models generating all four modalities, metrics are pooled across T1w, T2w, T1Gd, and FLAIR, whereas metrics for T1w-only external references are computed only on generated T1w samples. 
FID and MMD measure distributional alignment in MedicalNet feature space, where lower values indicate closer agreement with the real BrainScape distribution. 
MS-SSIM is computed between generated samples as a diversity metric. 
Lower values indicate lower self-similarity and therefore greater sample diversity.
\end{tablenotes}
\end{threeparttable}
\end{table}

The diffusion results for each modality are provided in Supplementary Material Appendix~B.
These results show that the performance of WFDM is contrast-dependent, with the strongest gains observed for T1w and T2w and smaller improvements for FLAIR.
For T1w, WFDM reduced FID from $0.00335228$ to $0.00159097$ and MMD from $0.183375$ to $0.123779$ relative to the internal MAISI baseline. 
For T2w, WFDM improved FID from $0.0126402$ to $0.00754393$ and MMD from $0.266439$ to $0.218096$. 
FLAIR results were more closely matched, but WFDM still achieved the strongest FID and MMD.

T1Gd remained the most challenging modality in the present setting. 
For this contrast, the internal MAISI baseline achieved lower FID and MMD, whereas WFDM attained slightly lower MS-SSIM between generated samples. 
This weaker T1Gd synthesis performance is broadly consistent with the comparatively weaker T1Gd reconstruction performance of the Wavelet-Fusion latent compressor, which may contribute to the remaining gap for this modality.

Supplementary Material Appendix~C provides qualitative comparisons across axial, sagittal, and coronal views for T1w, T2w, T1Gd, and FLAIR. 
These supplementary figures show that the strongest quantitative models also remain the most visually plausible across axial, sagittal, and coronal views, while again highlighting that T1Gd is the hardest contrast for the proposed WFDM generator.

\subsection{Scheduler and sampling ablation}

Table~\ref{tab:scheduler-pooled} reports the pooled solver and scheduler ablation study for the WFDM model family. 
RFlow with $v$-prediction and $\ell_2$ loss achieved the strongest overall realism, obtaining the lowest pooled FID and MMD while requiring only $50$ inference steps. 
Relative to the $1000$-step DDPM variants, RFlow with $50$ inference steps reduced the inference sampling steps by a factor of $20$ while also improving the main realism metrics (FID and MMD). 
Within DDPM, the cosine schedule consistently outperformed the scaled-linear alternative on FID and MMD. 
Although the scaled-linear schedule produced the lowest MS-SSIM among generated samples, this diversity gain was accompanied by substantially worse FID and MMD scores. 
Lower self-similarity alone is insufficient evidence of better generation quality when distributional alignment with real images deteriorates.
Overall, RFlow provided the best balance between realism, diversity, and sampling efficiency, and was therefore used as the default WFDM sampler.

\begin{table}[t]
\centering
\scriptsize
\begin{threeparttable}
\caption{Scheduler and sampler comparison pooled across modalities for the WFDM model family.}
\label{tab:scheduler-pooled}
\setlength{\tabcolsep}{7pt}%
\renewcommand{\arraystretch}{1.1}
\begin{tabular}{@{}lcccccc@{}}
\toprule
Scheduler / Sampler & Pred. type & Loss & Inference steps & Overall FID $\downarrow$ & Overall MMD $\downarrow$ & Overall MS-SSIM $\downarrow$ \\
\midrule
DDPM (cosine $\beta$ schedule) & $v$ & $\ell_1$ & 1000 & 0.00535477 & 0.184523 & 0.761192 \\
DDPM (cosine $\beta$ schedule) & $v$ & $\ell_2$ & 1000 & 0.00567393 & 0.163462 & 0.742717 \\
DDPM (scaled-linear $\beta$ schedule) & $v$ & $\ell_2$ & 1000 & 0.05349090 & 0.337763 & 0.436904 \\
RFlow & $v$ & $\ell_2$ & 50 & 0.00404259 & 0.147064 & 0.754384 \\
\bottomrule
\end{tabular}
\end{threeparttable}
\end{table}

\subsection{Inference-time efficiency results}

Table~\ref{tab:inference-main} summarizes the inference-time benchmark across all evaluated generators. 
The primary metric is latency per generated 3D MRI volume, measured around the model inference call and excluding data loading and file writing. 
Peak GPU memory is reported to reflect the practical resource requirements of each model under the same deployment setting.

Across the evaluated diffusion models, runtime depends on the number of denoising steps and on the synthesis domain in which generation is performed. 
WFDM generated a full $(192,224,160)$ volume in $2.161$ seconds on average using $50$ RFlow steps, compared with $1.919$ seconds for the MAISI baseline. 
The T1w-only LDM reference required $7.453$ seconds, WDM required $24.128$ seconds for a smaller $(128,128,128)$ output, and Med-DDPM required $95.86$ seconds for a $(192,192,144)$ MRI output. 
HA-GAN was the fastest model at $0.034$ seconds per sample because it does not require iterative denoising, but its quantitative realism metrics were substantially weaker than those of the latent diffusion models trained on BrainScape.

Table~\ref{tab:inference-wfdm-ablation} reports the corresponding efficiency benchmark for the WFDM sampler ablation study. 
This table isolates the effect of replacing $1000$-step DDPM inference with $50$-step RFlow inference while holding the remaining WFDM model family fixed. 
The three DDPM configurations required $26.849$ to $28.875$ seconds per generated volume on average, whereas RFlow required only $2.161$ seconds. 
This corresponds to a $12.4$ to $13.4$ times reduction in measured latency, while also achieving the lowest pooled FID and MMD in Table~\ref{tab:scheduler-pooled}. 
Together, the quality and runtime results indicate that RFlow is the most practical WFDM sampling configuration, combining the strongest distributional alignment with substantially faster inference.

\begin{table}[t]
\centering
\begin{threeparttable}
\caption{Inference-time benchmark of evaluated 3D synthetic MRI generators.\tnote{$a$}}
\label{tab:inference-main}
\setlength{\tabcolsep}{10pt}
\renewcommand{\arraystretch}{1.12}
\begin{tabular}{@{}lccrrrr@{}}
\toprule
Model & Steps & Output size & Mean & Median & Std & Peak mem. \\
      &       &             & s/vol $\downarrow$ & s/vol $\downarrow$ & s/vol $\downarrow$ & GB $\downarrow$ \\
\midrule
WFDM & 50 & $(192,224,160)$ & 2.161 & 2.106 & 0.0871 & 3.002 \\
MAISI & 50 & $(192,224,160)$ & 1.919 & 1.939 & 0.0855 & 14.370 \\
LDM & 50 & $(192,224,160)$ & 7.453 & 7.453 & 0.0241 & 13.470 \\
WDM & 1000 & $(128,128,128)$ & 24.128 & 24.183 & 0.2439 & 0.865 \\
Med-DDPM & 250 & $(192,192,144)$ & 95.860 & 95.890 & 0.1193 & 13.320 \\
HA-GAN & N/A & $(192,224,160)$ & 0.034 & 0.034 & 0.0004 & 4.769 \\
\bottomrule
\end{tabular}
\begin{tablenotes}[flushleft]\footnotesize
\item[${a}$]
Inference-time benchmark performed on a single NVIDIA H100 GPU on NeSI with batch size 1. 
Latency is reported in seconds per generated volume and excludes data loading and file writing. 
Each configuration was measured over 30 timed runs after 3 warm-up runs.
\textit{Steps} denotes the number of iterative sampling steps. N/A indicates a non-iterative GAN-based generator. 
\textit{Output size} reports the generated 3D volume dimensions, and \textit{Peak mem.} reports peak GPU memory usage in GB. 
Lower values indicate faster inference or lower memory use.
\end{tablenotes}
\end{threeparttable}
\end{table}

\begin{table}[t]
\centering
\begin{threeparttable}
\caption{Inference-time benchmark of WFDM under different sampler settings.\tnote{$a$}}
\label{tab:inference-wfdm-ablation}
\setlength{\tabcolsep}{8pt}%
\renewcommand{\arraystretch}{1.12}
\begin{tabular}{@{}lccrrrrr@{}}
\toprule
Sampler & Pred. type & Loss & Steps & Mean & Median & Std & Peak mem. \\
        &            &      &       & s/vol $\downarrow$ & s/vol $\downarrow$ & s/vol $\downarrow$ & GB $\downarrow$ \\
\midrule
DDPM (cosine $\beta$) & $v$ & $\ell_1$ & 1000 & 28.875 & 28.844 & 1.2251 & 3.002 \\
DDPM (cosine $\beta$) & $v$ & $\ell_2$ & 1000 & 28.194 & 28.448 & 1.4040 & 3.002 \\
DDPM (scaled-linear $\beta$) & $v$ & $\ell_2$ & 1000 & 26.849 & 26.603 & 0.5813 & 3.002 \\
RFlow & $v$ & $\ell_2$ & 50 & 2.161 & 2.106 & 0.0871 & 3.002 \\
\bottomrule
\end{tabular}
\begin{tablenotes}[flushleft]\footnotesize
\item[${a}$]
WFDM sampler settings are benchmarked using the same protocol as Table~\ref{tab:inference-main}, with output size $(192,224,160)$. 
\textit{Pred. type}, \textit{Loss}, and \textit{Steps} denote the prediction target, training loss objective, and number of inference sampling steps, respectively. 
Lower values indicate faster inference or lower peak GPU memory use.
\end{tablenotes}
\end{threeparttable}
\end{table}

\FloatBarrier
\section{Discussion}

The results of this study show that large-scale, multimodal 3D brain MRI synthesis is feasible with 3D latent diffusion models trained on a heterogeneous pooled dataset such as BrainScape.  
Our proposed WFDM achieved the strongest pooled distributional alignment to real BrainScape data among the evaluated generators, while maintaining comparable sample diversity.  
A likely explanation for this advantage is the Wavelet-Fusion latent representation used by WFDM.  
In WFDM, the denoising model operates in a latent space where compression is guided by both learned convolutional features and spatial-frequency information. 
This matters because the quality of latent diffusion is constrained by the information preserved during first-stage compression \parencite{rombach-LDM-2022}.  
By preserving low-frequency anatomical structure together with higher-frequency information related to tissue boundaries, ventricular margins, and contrast transitions, the WF-SA-VAE compressor provides a compact but more anatomically informative latent representation for synthesis \parencite{Friedrich-WDM-2024}. 
This interpretation is consistent with the compressor benchmark, where WF-SA-VAE was the strongest BrainScape-trained compressor, and with the synthesis benchmark, where WFDM achieved the lowest pooled FID and MMD among the evaluated generators. 
Furthermore, the results for each modality suggest that a single conditional generator can capture anatomical structure shared across structural MRI modalities while producing 3D MRI volumes with the requested contrast when the target modality is provided explicitly as conditioning.  
These findings support the central motivation of the paper by showing that a controllable latent diffusion model trained on the heterogeneous BrainScape dataset can synthesize multimodal brain MRI across T1w, T2w, T1Gd, and FLAIR.

A key scientific and clinical implication of this work is that synthetic MRI can help researchers address uneven modality availability in large pooled neuroimaging resources. 
In BrainScape and similar public neuroimaging collections, T1w MRI is widely represented, whereas T2w, FLAIR, and especially T1Gd are less consistently available across subjects and source datasets \parencite{Yasinzai-BrainScape-2025}. 
The value of WFDM is therefore not simply that it can generate biologically plausible images, but it also provides a controllable synthesis framework for producing specified target modalities within a heterogeneous multi-study setting.
In its present form, WFDM supports synthetic cohort generation and dataset augmentation conditioned on the requested target modality and available metadata, rather than synthesizing a missing contrast from the available MRI scans of the same subject. 
Subject-specific modality completion would require an additional image-conditioning pathway, such as a ControlNet adapter, so that available MRI contrasts from an individual subject can provide anatomical guidance for generating the missing target modality \parencite{zhang-ControlNet-2023}. 
Thus, the current model provides a foundation for controlled multimodal cohort synthesis, while future image-conditioned extensions could support subject-specific modality completion workflows \parencite{Koetzier-SyntheticDataReview-2024,Kim-ALDM-2024,Meng-FLAIRSynth-2024,zhang-ControlNet-2023}.

Furthermore, our results indicate that downstream diffusion quality is constrained by the quality of the first-stage compressor. 
The pooled compressor results in Table~\ref{tab:ae-pooled} show that the Wavelet-Fusion autoencoders trained on BrainScape substantially outperform the other alternatives trained on BrainScape, with WF-SA-VAE achieving the strongest pooled reconstruction among the learned compressors trained on BrainScape. 
Although the externally trained MAISI VAE-GAN achieved the strongest overall reconstruction metrics in the compressor benchmark, the downstream diffusion results show that WF-SA-VAE provides an effective latent space for multimodal synthesis on BrainScape. 
In practice, when the compressor removes fine anatomical detail or introduces reconstruction artifacts, the downstream denoising model is constrained by those information losses.
This pattern is also visible at the modality level: T1Gd is the weakest reconstruction case for WF-SA-VAE, with poorer reconstruction fidelity than T1w, T2w, and FLAIR, and it is likewise the most challenging modality in the downstream diffusion results, where WFDM remains weaker than the MAISI baseline on FID and MMD. 
This interpretation is consistent with prior latent diffusion work showing that the first-stage autoencoder is not merely a dimensionality-reduction module. 
It is a critical component that determines the balance between compression efficiency and information preservation. 
The latent representation must be compact enough to make high-resolution 3D diffusion computationally practical, while retaining sufficient anatomical detail, perceptual fidelity, and latent regularization for the diffusion model to learn the target image distribution effectively \parencite{rombach-LDM-2022,Pinaya-LDM3DBrain-2022,Guo-Maisi-2025}.

This study also introduces demographic and clinical metadata conditioning with an explicit metadata availability mask for multimodal MRI synthesis. 
Rather than conditioning on raw metadata alone, the model uses a schema with a fixed field order to convert each subject's metadata into a structured numeric vector and a corresponding metadata availability mask that indicates which fields are available.
These representations are encoded by a dedicated metadata encoder and injected into the 3D latent diffusion model. 
This design is important for BrainScape because metadata coverage is broad but uneven across source datasets, with some variables available for many subjects and others recorded only in specific source datasets or absent entirely \parencite{Yasinzai-BrainScape-2025}. 
This metadata availability mask allows the model to distinguish observed covariates from unavailable fields, rather than treating missing values as explicit negative or zero-valued entries.
This distinction is especially important for clinical and pathology indicators, which are available only for subsets of BrainScape source datasets and should be interpreted as source-dataset conditioning labels rather than complete annotations across the combined BrainScape dataset.
This allows metadata conditioning to be used in a heterogeneous pooled dataset without requiring every contributing dataset to provide the same demographic or clinical variables.


Evaluation methodology for synthetic 3D MRI is not yet fully standardized. 
Quantitative evaluation results depend on the selected metric, preprocessing and normalization steps, the feature space used for distributional comparisons, and whether evaluation is based on full 3D volumes or slice-based representations.
As a result, direct comparison across studies remains difficult unless the evaluation protocol, feature extractor, preprocessing pipeline, and normalization strategy are clearly specified \parencite{Dohmen-SynthEvalMetrics-2025,Koetzier-SyntheticDataReview-2024}.
To provide a more consistent and reproducible evaluation framework for synthetic 3D MRI, we release and use the \textit{medmetric} package, which computes FID and MMD in a 3D MedicalNet feature space while reporting MS-SSIM separately in image space as a diversity proxy \parencite{Yasinzai-MedMetricGitHub}. 
This makes the feature extractor and preprocessing assumptions explicit, which is important because FID and MMD values depend on the feature space in which real and synthetic images are compared.

Participant privacy, consent, and data governance are important considerations when working with neuroimaging data. 
Synthetic MRI can support dataset augmentation, modality completion, and controlled synthetic cohort generation, but synthetic data are not automatically privacy-preserving. 
Generative models can memorize training examples or leak information about the training distribution, so synthetic medical images should not be assumed safe for unrestricted sharing without appropriate privacy evaluation \parencite{Akbar-BewareDiffusionMemorization-2025,Daum-DP3DLatentDiffusion-2024,Shi-SynthDataPrivacy-2025}. 
WFDM is trained on BrainScape, a resource derived from publicly available MRI source datasets, placing this work in a public MRI data synthesis setting \parencite{Yasinzai-BrainScape-2025}.

A further consideration is the risk of model collapse when synthetic samples are recursively used to train future generative models, shifting the training distribution away from the original real-data distribution.
Model collapse is a degenerative process in which each successive generative model learns increasingly from the outputs of previous models rather than from the underlying real distribution, causing the learned distribution to drift over generations \parencite{shumailov-ModelCollapse-2024}.
Across recursive generations, sampling and modelling errors can accumulate, causing low probability cases or features to disappear first, while later generations may represent the original distribution with reduced diversity or increasing distortion \parencite{shumailov-ModelCollapse-2024}. 
This concern is especially relevant in medical imaging, where rare diseases, minority subgroups, uncommon acquisition protocols, and underrepresented MRI contrasts often occupy the tails of the data distribution.
These are precisely the cases that clinical AI systems must preserve rather than lose. 
Therefore, images generated by WFDM should be treated as labelled synthetic supplements to real MRI data, not as untracked substitutes for real scans.
Practical safeguards therefore include preserving the provenance of synthetic samples, clearly distinguishing generated images from real scans, and ensuring that future generative training remains anchored to real MRI data rather than relying recursively on synthetic outputs alone.

The current framework can benefit neuroimaging researchers working with incomplete, heterogeneous, or imbalanced MRI datasets. 
By enabling controlled synthesis under specified modality and metadata conditions, WFDM provides a way to investigate how modality availability, cohort composition, and acquisition heterogeneity may affect downstream model performance and robustness. 
This is especially relevant for underrepresented MRI contrasts, small clinical cohorts, and rare neurological conditions, where real datasets are often limited, fragmented across studies, or acquired with inconsistent protocols. 
At the same time, these are also the cases where generated samples require careful interpretation, because limited representation in the training data constrains what the model can learn.

\subsection{Limitations}
A first limitation of this study is that most available 3D brain MRI synthesis baselines are not directly designed for the multimodal MRI generation setting addressed in this work. 
WFDM is trained as a single controllable diffusion model that can generate four target MRI modalities, T1w, T2w, T1Gd, and FLAIR, using explicit modality conditioning. 
Among the evaluated baselines, the MAISI model provides the closest matched comparison because it can be modified to use the requested MRI modality as conditioning and generate different target contrasts. 
In contrast, Med-DDPM is designed for mask-conditioned brain MRI synthesis and generates multimodal tumor MRI from anatomical masks rather than explicit target-modality conditioning \parencite{Dorjsembe-MedDDPM-2024}. 
Adapting Med-DDPM to BrainScape would therefore require substantial changes, including replacing mask conditioning with target-modality conditioning and retraining the model on BrainScape. 
Retraining Med-DDPM on BrainScape would substantially increase training time, computational cost, and implementation complexity, given BrainScape's larger dataset size, greater cross-dataset heterogeneity, and uneven modality availability across subjects \parencite{Yasinzai-BrainScape-2025}.
Similarly, pretrained generators configured for a single output modality, such as LDM and HA-GAN, provide useful external references, but their released configurations are limited to T1w synthesis and do not support unified generation of T1w, T2w, T1Gd, and FLAIR \parencite{Pinaya-LDM3DBrain-2022,Sun-HAGAN-2022}.
Adapting these models to the BrainScape multimodal setting would require either training separate models for each modality or modifying the architectures to support explicit target-modality conditioning. 
Both options would introduce additional feasibility challenges, including increased implementation effort, substantially higher computational cost, and longer training time. 
For these reasons, the MRI synthesis comparisons in this study are meaningful, but not fully exhaustive.

A second limitation is that although WFDM incorporates metadata conditioning, the present study focuses on overall synthesis quality and does not include a dedicated evaluation of metadata controllability. 
Specifically, the model includes a metadata encoder that takes both the encoded metadata vector and the corresponding metadata availability mask as input, allowing the conditioning pathway to distinguish observed covariate values from unavailable fields. 
However, this paper does not systematically evaluate whether changing individual metadata fields produces consistent and expected changes in the generated images. 
This evaluation is therefore an important direction for future work, while the present study focuses on establishing the core multimodal synthesis framework.

A third limitation is that the current WFDM model supports controllable synthetic MRI generation, but does not yet perform modality completion for individual subjects.
At its current stage, WFDM can generate any of the four target modalities by conditioning on the requested modality and available metadata. 
However, it does not yet condition directly on the available MRI modalities from the same subject to synthesize that subject's missing contrasts. 
Modality completion for individual subjects requires an additional conditioning pathway that uses the available MRI modalities from the same subject to preserve individual anatomy while generating the missing target modality.

Finally, T1Gd remains the most challenging modality for the WFDM model. 
This likely reflects both the smaller number of T1Gd scans in the BrainScape training subset and the comparatively weaker T1Gd reconstruction performance of the proposed latent compressor. 
This result does not undermine the overall contribution of WFDM, because the model still supports unified generation across T1w, T2w, T1Gd, and FLAIR and achieves the strongest pooled distributional alignment among the evaluated synthetic MRI generators. 
However, it highlights room for further improvement, including more balanced sampling across modalities, additional T1Gd data, and improved latent representation for rare modalities.

\subsection{Future work}
This study is a step toward our broader objective of using synthetic MRI for downstream multimodal neuroimaging and clinical AI applications
The next direction is to extend WFDM from controllable multimodal synthesis to subject-specific modality completion. 
The current WFDM model already provides two key components for this extension. 
First, conditioning on the target modality specifies the MRI contrast to be generated. 
Second, metadata conditioning, together with the metadata availability mask, incorporates available demographic and clinical context.
Future work will add an image-conditioning adapter, for example a ControlNet conditioning pathway \parencite{zhang-ControlNet-2023}, on top of the trained WFDM backbone so that available MRI modalities from an individual subject can guide the generation of missing target modalities. 
This design allows the trained WFDM backbone to be reused for multimodal MRI generation, while the additional image-conditioning adapter provides subject-specific anatomical guidance from the available source modalities and directs generation toward the missing target contrast, without requiring the entire diffusion model to be retrained from scratch. 
This would enable a completed BrainScape resource in which each subject can be represented across T1w, T2w, T1Gd, and FLAIR using the originally available scans together with synthesized missing contrasts where appropriate.
Such a completed multimodal dataset would support downstream studies that require complete modality sets, including brain tumor segmentation and other clinical AI tasks.

Another future direction is to perform a dedicated evaluation of metadata controllability. 
Although the present study introduces metadata conditioning that explicitly represents unavailable fields, future work should assess whether WFDM uses this conditioning pathway in a consistent and interpretable manner.
This could include controlled generation experiments, subgroup-specific analyses, and sensitivity tests to assess whether available demographic or clinical covariates influence generated images in consistent and interpretable ways.

Future work will also include a broader benchmark of baseline synthesis models trained on BrainScape, where feasible.
The present study includes baselines trained on BrainScape and pretrained external references, but a fully standardized benchmark would require retraining additional baseline architectures directly on BrainScape. 
Where architectural adaptation, training time, and computational cost are feasible, future work will extend this comparison by retraining additional models in the BrainScape setting.

Together, these directions will move WFDM beyond multimodal MRI synthesis toward subject-specific modality completion, enabling future downstream AI applications that require complete multimodal MRI inputs.

\FloatBarrier
\section{Conclusion}

In this study, we present WFDM, a controllable multimodal 3D latent diffusion framework for brain MRI synthesis trained and evaluated on BrainScape. 
The model generates T1w, T2w, T1Gd, and FLAIR volumes through explicit target-modality conditioning, while also supporting metadata conditioning with an explicit metadata availability mask. 
By building on BrainScape, this work addresses the practical challenge of uneven modality availability in large pooled neuroimaging datasets and provides a unified generative framework for synthesizing multiple structural MRI contrasts across heterogeneous public data.

In pooled BrainScape evaluation, WFDM achieved the strongest overall distributional alignment among the evaluated synthetic MRI generators while maintaining comparable synthetic-sample diversity. 
Modality-specific analysis showed improvements for T1w, T2w, and FLAIR synthesis, whereas T1Gd remained the most challenging contrast, with the MAISI baseline achieving lower FID and MMD for this modality. 
The sampler ablation further showed that RFlow inference with 50 sampling steps achieved strong distributional realism at substantially lower sampling cost than 1000-step DDPM variants, supporting the practical use of WFDM for large-scale synthetic MRI generation.

A central component of this framework is the Wavelet-Fusion VAE, which provides the latent compression stage for diffusion. 
This component improves the quality of the learned latent representation compared with other compressor baselines trained on BrainScape and supports downstream multimodal diffusion synthesis by preserving anatomical detail within a compact latent space. 
Together, the Wavelet-Fusion latent representation, target-modality conditioning, and metadata conditioning with a metadata availability mask enable WFDM to synthesize controllable multimodal brain MRI across the heterogeneous BrainScape dataset.

Overall, these findings show that WFDM can generate controllable multimodal brain MRI and provides a foundation for dataset augmentation, controlled synthetic cohort generation, and future modality completion workflows. 
In future work, WFDM can be extended with additional image-conditioning adapters so that available subject-specific MRI modalities can guide the synthesis of missing contrasts. 
This would support the creation of completed multimodal datasets in which each subject could have a combination of real and synthetic modalities. 
Future work should also focus on subject-specific image-conditioned modality completion and on downstream validation in clinical AI tasks that require complete multimodal MRI inputs.

\FloatBarrier
\section{Data availability} 

This study uses BrainScape, an open-source framework that aggregates public brain MRI datasets through dataset-specific configuration files and metadata \parencite{Yasinzai-BrainScape-2025}. 
Researchers should refer to the BrainScape paper for details on data access, dataset reconstruction, and the preprocessing workflow used to generate the study data from the original public sources. 

\FloatBarrier
\section{Code availability} 

The code for this work is available at \url{http://github.com/yasinzaii/generative-mri}. 
The repository includes the BrainScape data-handling components used in this study, the complete training pipeline, model configuration files, inference and sampling scripts, and the environment specification required to reproduce the experiments reported in this paper. 
The evaluation scripts are also included and use \texttt{medmetric}, which provides the metric implementations used for synthetic MRI evaluation in this work.

\FloatBarrier
\section*{Author Contributions}

\begin{enumerate}
  \item \textbf{Muhammad Nabi Yasinzai (MNY)}: Conceptualization, Methodology, Software, Writing - Original Draft.
  \item \textbf{Remika Mito (RM)}: Conceptualization, Supervision, Writing - Review \& Editing. 
  \item \textbf{Mangor Pedersen (MP)} Conceptualization, Supervision, Writing - Review \& Editing, Resources, Funding acquisition. 
\end{enumerate}
All authors read, revised, and approved the final version of the manuscript.

\section*{Funding}

MNY is a recipient of Australian Epilepsy Project PhD scholarship 
(Medical Research Future Fund - Frontier Health and Medical Research Program - 
Grant Numbers MRFF75908). 
RM is the recipient of an Australian Research Council Discovery Early Career 
Researcher Award (project number: DE240101035). 
MP is a recipient of Health Research Council fellowship, New Zealand (\#21/622).

\section*{Declaration of Competing Interests}

The authors declare that they have no competing financial or non-financial interests 
that could have influenced the work reported in this manuscript.

\section*{Acknowledgements}

We thank the participants in the original studies whose data make this work possible. 
Their contributions to open neuroimaging resources form the foundation of BrainScape's 
mission to build a broad, diverse, and evolving dataset.

We also acknowledge New Zealand eScience Infrastructure (NeSI) for providing the high-performance computing resources that supported the training and evaluation of the models in this study.

\FloatBarrier
\section{Supplementary Material}

\FloatBarrier
\subsection{Appendix A: Modality-specific autoencoder reconstruction results}

This appendix reports the full modality-specific autoencoder reconstruction results corresponding to the pooled analysis in the main paper. 
Tables~\ref{tab:ae-t1w}--\ref{tab:ae-flair} provide LPIPS, SSIM, MS-SSIM, PSNR, and MSE separately for T1w, T2w, T1Gd, and FLAIR. 
These tables are included to verify that the pooled trends reported in the main text remain stable across individual MRI contrasts and are not driven by a single modality. 
They also make it possible to inspect modality-specific reconstruction difficulty and to compare how consistently each latent compressor performs across the full BrainScape setting. 

Consistent with the pooled comparison, WF-SA-VAE was the strongest learned compressor trained on BrainScape in every modality. 
WF-VAE remained the second strongest BrainScape-trained variant, indicating that fusing wavelet features within the autoencoder contributes substantially to reconstruction quality, while the additional self-attention block provides a further consistent gain. 
By contrast, the simpler Wavelet VAE, which uses the two-stage wavelet decomposition as input to a simplified autoencoder, performed notably worse than the variants that fuse wavelet features within the autoencoder across all four modalities. 

The external MAISI VAE-GAN reference remained the strongest learned compressor overall in each modality. 
However, the gap between MAISI VAE-GAN and WF-SA-VAE was relatively small for T1w, T2w, and FLAIR, whereas T1Gd showed the clearest separation and was the most difficult modality overall. 
This pattern is reflected in the lower SSIM and higher LPIPS and MSE values observed for T1Gd across nearly all learned models. 
The non-learned one-stage and two-stage wavelet transforms are also reported as nearly lossless transform references rather than practical learned compression models. 

\begin{table}[bp]
\centering
\scriptsize
\begin{threeparttable}
\caption{Autoencoder reconstruction metrics for T1w.}
\label{tab:ae-t1w}
\setlength{\tabcolsep}{6pt}
\renewcommand{\arraystretch}{1.1}
\begin{tabular}{@{}lccccccc@{}}
\toprule
Model & Compression & Latent chans & LPIPS $\downarrow$ & SSIM $\uparrow$ & MS-SSIM $\uparrow$ & PSNR $\uparrow$ & MSE $\downarrow$ \\
\midrule
ALDM VQ-GAN & -- & -- & 0.4675 & 0.8070 & 0.8648 & 17.26 & 0.549055 \\
BrainSynth VQ-VAE & -- & -- & 0.6120 & 0.6654 & 0.8112 & 19.65 & 0.011148 \\
LDM-VAE & 1/8 & 3 & 0.2347 & 0.8714 & 0.9590 & 23.01 & 0.006340 \\
VAE-GAN (BrainScape) & 1/4 & 4 & 0.1438 & 0.6023 & 0.9839 & 27.40 & 0.002331 \\
MAISI VAE-GAN & 1/4 & 4 & 0.0388 & 0.9706 & 0.9961 & 32.90 & 0.000555 \\
Wavelet VAE & 1/4 & 4 & 0.2655 & 0.6015 & 0.9696 & 24.51 & 0.003663 \\
WF-VAE & 1/4 & 4 & 0.0724 & 0.9610 & 0.9933 & 29.71 & 0.001128 \\
WF-SA-VAE & 1/4 & 4 & 0.0576 & 0.9676 & 0.9953 & 30.95 & 0.000840 \\
WDM-style DWT (1-stage) & 1/2 & 8 & 0.0000 & 1.0000 & 1.0000 & 82.04 & 0.000000 \\
DWT reference (2-stage) & 1/4 & 64 & 0.0000 & 1.0000 & 1.0000 & 76.02 & 0.000000 \\
\bottomrule
\end{tabular}
\end{threeparttable}
\end{table}

\begin{table}[t]
\centering
\scriptsize
\begin{threeparttable}
\caption{Autoencoder reconstruction metrics for T2w.}
\label{tab:ae-t2w}
\setlength{\tabcolsep}{6pt}
\renewcommand{\arraystretch}{1.1}
\begin{tabular}{@{}lccccccc@{}}
\toprule
Model & Compression & Latent chans & LPIPS $\downarrow$ & SSIM $\uparrow$ & MS-SSIM $\uparrow$ & PSNR $\uparrow$ & MSE $\downarrow$ \\
\midrule
ALDM VQ-GAN & -- & -- & 0.5017 & 0.7659 & 0.6938 & 15.58 & 0.390046 \\
BrainSynth VQ-VAE & -- & -- & 0.6175 & 0.6639 & 0.8097 & 20.55 & 0.009422 \\
LDM-VAE & 1/8 & 3 & 0.2179 & 0.8579 & 0.9471 & 23.07 & 0.006680 \\
VAE-GAN (BrainScape) & 1/4 & 4 & 0.1177 & 0.6946 & 0.9845 & 28.06 & 0.002395 \\
MAISI VAE-GAN & 1/4 & 4 & 0.0408 & 0.9698 & 0.9951 & 32.35 & 0.000649 \\
Wavelet VAE & 1/4 & 4 & 0.3383 & 0.7066 & 0.9687 & 24.69 & 0.003652 \\
WF-VAE & 1/4 & 4 & 0.0836 & 0.9601 & 0.9928 & 30.00 & 0.001131 \\
WF-SA-VAE & 1/4 & 4 & 0.0668 & 0.9680 & 0.9949 & 31.15 & 0.000819 \\
WDM-style DWT (1-stage) & 1/2 & 8 & 0.0000 & 1.0000 & 1.0000 & 84.97 & 0.000000 \\
DWT reference (2-stage) & 1/4 & 64 & 0.0000 & 1.0000 & 1.0000 & 78.95 & 0.000000 \\
\bottomrule
\end{tabular}
\end{threeparttable}
\end{table}

\begin{table}[t]
\centering
\scriptsize
\begin{threeparttable}
\caption{Autoencoder reconstruction metrics for T1Gd.}
\label{tab:ae-t1ce}
\setlength{\tabcolsep}{6pt}
\renewcommand{\arraystretch}{1.1}
\begin{tabular}{@{}lccccccc@{}}
\toprule
Model & Compression & Latent chans & LPIPS $\downarrow$ & SSIM $\uparrow$ & MS-SSIM $\uparrow$ & PSNR $\uparrow$ & MSE $\downarrow$ \\
\midrule
ALDM VQ-GAN & -- & -- & 0.5051 & 0.7670 & 0.7377 & 15.77 & 0.390147 \\
BrainSynth VQ-VAE & -- & -- & 0.7167 & 0.6498 & 0.8006 & 20.18 & 0.011132 \\
LDM-VAE & 1/8 & 3 & 0.2442 & 0.8340 & 0.9358 & 21.94 & 0.008924 \\
VAE-GAN (BrainScape) & 1/4 & 4 & 0.1489 & 0.6338 & 0.9776 & 26.28 & 0.004171 \\
MAISI VAE-GAN & 1/4 & 4 & 0.0560 & 0.9463 & 0.9918 & 30.47 & 0.001058 \\
Wavelet VAE & 1/4 & 4 & 0.3853 & 0.6982 & 0.9564 & 23.38 & 0.005237 \\
WF-VAE & 1/4 & 4 & 0.1173 & 0.9332 & 0.9886 & 28.30 & 0.001813 \\
WF-SA-VAE & 1/4 & 4 & 0.0912 & 0.9396 & 0.9906 & 28.91 & 0.001349 \\
WDM-style DWT (1-stage) & 1/2 & 8 & 0.0000 & 1.0000 & 1.0000 & 82.97 & 0.000000 \\
DWT reference (2-stage) & 1/4 & 64 & 0.0000 & 1.0000 & 1.0000 & 76.94 & 0.000000 \\
\bottomrule
\end{tabular}
\end{threeparttable}
\end{table}

\begin{table}[t]
\centering
\scriptsize
\begin{threeparttable}
\caption{Autoencoder reconstruction metrics for FLAIR.}
\label{tab:ae-flair}
\setlength{\tabcolsep}{6pt}
\renewcommand{\arraystretch}{1.1}
\begin{tabular}{@{}lccccccc@{}}
\toprule
Model & Compression & Latent chans & LPIPS $\downarrow$ & SSIM $\uparrow$ & MS-SSIM $\uparrow$ & PSNR $\uparrow$ & MSE $\downarrow$ \\
\midrule
ALDM VQ-GAN & -- & -- & 0.4597 & 0.7941 & 0.8252 & 16.90 & 0.496206 \\
BrainSynth VQ-VAE & -- & -- & 0.6668 & 0.6575 & 0.8127 & 20.25 & 0.010016 \\
LDM-VAE & 1/8 & 3 & 0.2405 & 0.8511 & 0.9496 & 23.12 & 0.006415 \\
VAE-GAN (BrainScape) & 1/4 & 4 & 0.1565 & 0.5577 & 0.9785 & 26.62 & 0.003082 \\
MAISI VAE-GAN & 1/4 & 4 & 0.0483 & 0.9595 & 0.9937 & 32.06 & 0.000709 \\
Wavelet VAE & 1/4 & 4 & 0.3176 & 0.6253 & 0.9651 & 24.37 & 0.003910 \\
WF-VAE & 1/4 & 4 & 0.0928 & 0.9521 & 0.9915 & 29.51 & 0.001232 \\
WF-SA-VAE & 1/4 & 4 & 0.0737 & 0.9578 & 0.9931 & 30.40 & 0.000975 \\
WDM-style DWT (1-stage) & 1/2 & 8 & 0.0000 & 1.0000 & 1.0000 & 82.46 & 0.000000 \\
DWT reference (2-stage) & 1/4 & 64 & 0.0000 & 1.0000 & 1.0000 & 76.44 & 0.000000 \\
\bottomrule
\end{tabular}
\end{threeparttable}
\end{table}

\FloatBarrier
\subsection{Appendix B: Modality-specific diffusion synthesis results}

This appendix reports the modality-specific diffusion synthesis results. 
Tables~\ref{tab:diff-t1w}--\ref{tab:diff-flair} provide FID, MMD, and MS-SSIM separately for T1w, T2w, T1Gd, and FLAIR.
FID and MMD quantify distributional realism in MedicalNet feature space, where lower values indicate closer agreement with the real BrainScape distribution. 
Here, MS-SSIM is computed between generated samples as an image-space diversity proxy with lower values indicating lower self-similarity and therefore greater sample diversity.

Overall, the results for individual modalities are consistent with the pooled comparison, showing that WFDM performs best among the diffusion models trained on BrainScape in most cases.
For T1w and T2w, WFDM achieved clear gains, with substantially lower FID and MMD than the MAISI baseline. 
For FLAIR, WFDM also achieved lower FID and MMD, although the difference relative to the internal baseline was smaller. 
T1Gd remained the most challenging modality, with the internal MAISI baseline outperforming WFDM on FID and MMD scores. 
This contrast-dependent behavior is consistent with the reconstruction study, where T1Gd was also the most difficult modality for the learned latent compressors. 
The external pretrained baselines generally showed weaker distributional alignment relative to the baselines trained on BrainScape.

\begin{table}[t]
\centering
\scriptsize
\begin{threeparttable}
\caption{T1w synthesis metrics.}
\label{tab:diff-t1w}
\setlength{\tabcolsep}{10pt}
\renewcommand{\arraystretch}{1.1}
\begin{tabular}{@{}lccc@{}}
\toprule
Model & FID $\downarrow$ & MMD $\downarrow$ & MS-SSIM $\downarrow$ \\
\midrule
WFDM  & 0.00159097 & 0.123779 & 0.786674 \\
MAISI & 0.00335228 & 0.183375 & 0.792550 \\
LDM   & 0.01900610 & 0.411644 & 0.825484 \\
Med-DDPM  & 0.06895070 & 0.398040 & 0.602110 \\
WDM & 0.05860420 & 0.444065 & 0.733082 \\
HA-GAN  & 1.85993000 & 0.934294 & 0.694228 \\
\bottomrule
\end{tabular}
\end{threeparttable}
\end{table}

\begin{table}[t]
\centering
\scriptsize
\begin{threeparttable}
\caption{T2w synthesis metrics.}
\label{tab:diff-t2w}
\setlength{\tabcolsep}{10pt}
\renewcommand{\arraystretch}{1.1}
\begin{tabular}{@{}lccc@{}}
\toprule
Model & FID $\downarrow$ & MMD $\downarrow$ & MS-SSIM $\downarrow$ \\
\midrule
WFDM & 0.00754393 & 0.218096 & 0.677926 \\
MAISI & 0.01264020 & 0.266439 & 0.658323 \\
Med-DDPM & 0.02399400 & 0.247928 & 0.559383 \\
\bottomrule
\end{tabular}
\end{threeparttable}
\end{table}

\begin{table}[t]
\centering
\scriptsize
\begin{threeparttable}
\caption{T1Gd synthesis metrics.}
\label{tab:diff-t1ce}
\setlength{\tabcolsep}{10pt}
\renewcommand{\arraystretch}{1.1}
\begin{tabular}{@{}lccc@{}}
\toprule
Model & FID $\downarrow$ & MMD $\downarrow$ & MS-SSIM $\downarrow$ \\
\midrule
WFDM & 0.0241885 & 0.289871 & 0.740414 \\
MAISI & 0.0151833 & 0.233578 & 0.751685 \\
Med-DDPM & 0.0483055 & 0.272339 & 0.617553 \\
\bottomrule
\end{tabular}
\end{threeparttable}
\end{table}

\begin{table}[t]
\centering
\scriptsize
\begin{threeparttable}
\caption{FLAIR synthesis metrics.}
\label{tab:diff-flair}
\setlength{\tabcolsep}{10pt}
\renewcommand{\arraystretch}{1.1}
\begin{tabular}{@{}lccc@{}}
\toprule
Model & FID $\downarrow$ & MMD $\downarrow$ & MS-SSIM $\downarrow$ \\
\midrule
WFDM & 0.00324606 & 0.116996 & 0.705553 \\
MAISI & 0.00336307 & 0.132238 & 0.719311 \\
Med-DDPM & 0.01801030 & 0.231053 & 0.582306 \\
\bottomrule
\end{tabular}
\end{threeparttable}
\end{table}

\FloatBarrier
\subsection{Appendix C: Qualitative diffusion synthesis comparisons}

\begin{figure*}[t]
  \centering
  \includegraphics[width=\textwidth]{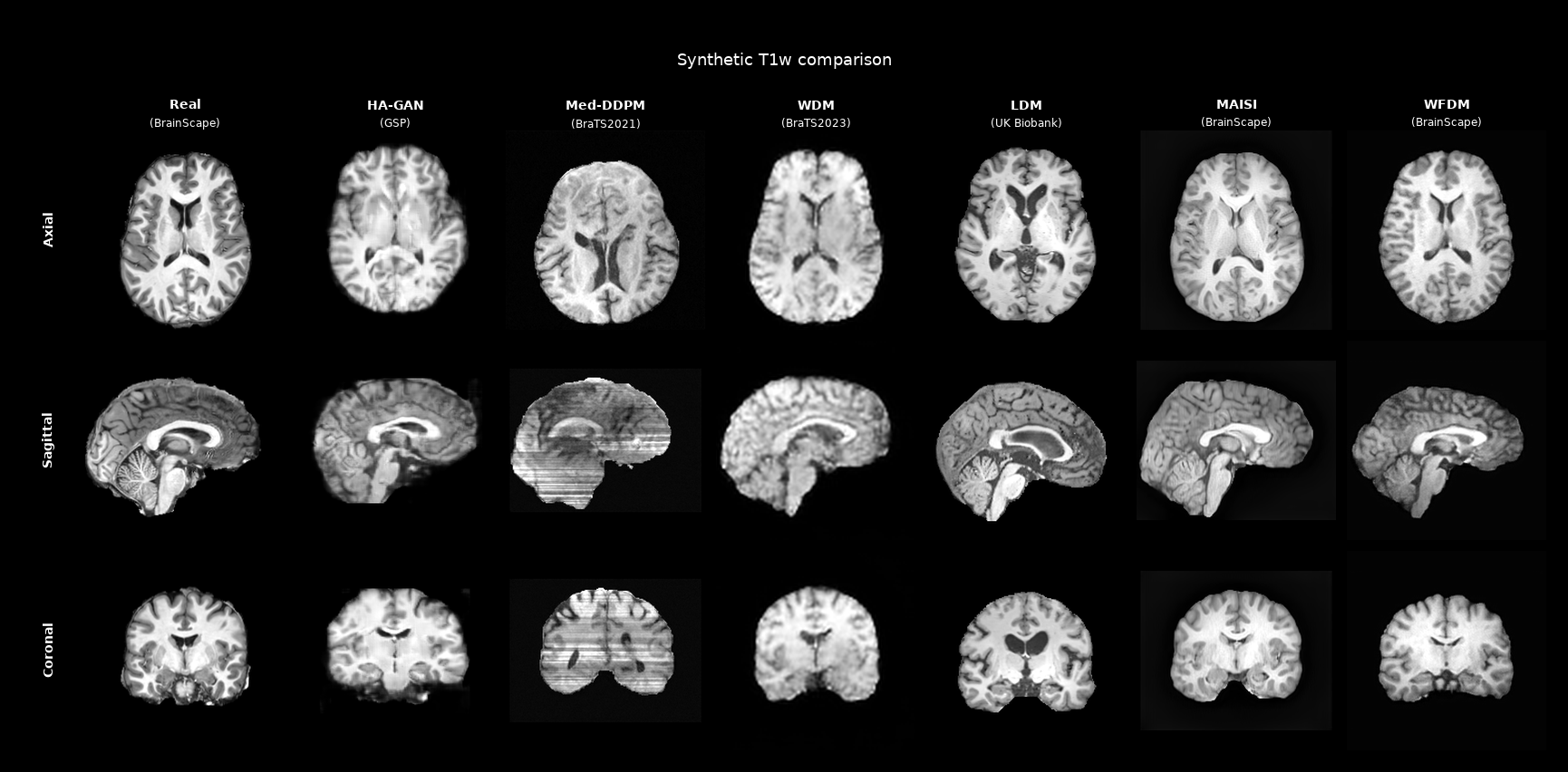}
  \caption{Qualitative comparison for T1w synthesis. A real BrainScape MRI is compared with synthetic T1w MRI sampled via HA-GAN (GSP), Med-DDPM (BraTS2021), WDM (BraTS2023), LDM (UK Biobank), MAISI (BrainScape), and WFDM (BrainScape) across axial, sagittal, and coronal views.}
  \label{fig:appendix-diff-t1w}
\end{figure*}

\begin{figure*}[t]
  \centering
  \includegraphics[width=0.82\textwidth]{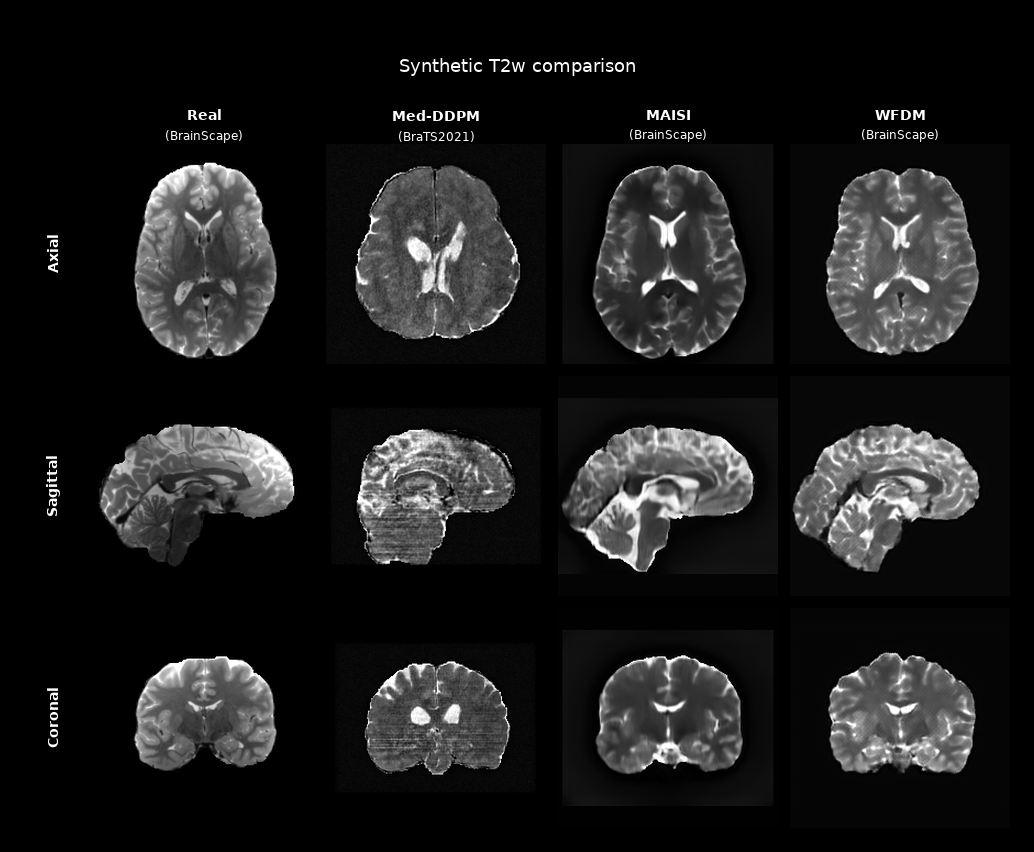}
  \caption{Qualitative comparison for T2w synthesis. The real BrainScape target image is compared with Med-DDPM (BraTS2021), MAISI (BrainScape), and WFDM (BrainScape) across axial, sagittal, and coronal views.}
  \label{fig:appendix-diff-t2w}
\end{figure*}

\begin{figure*}[t]
  \centering
  \includegraphics[width=0.82\textwidth]{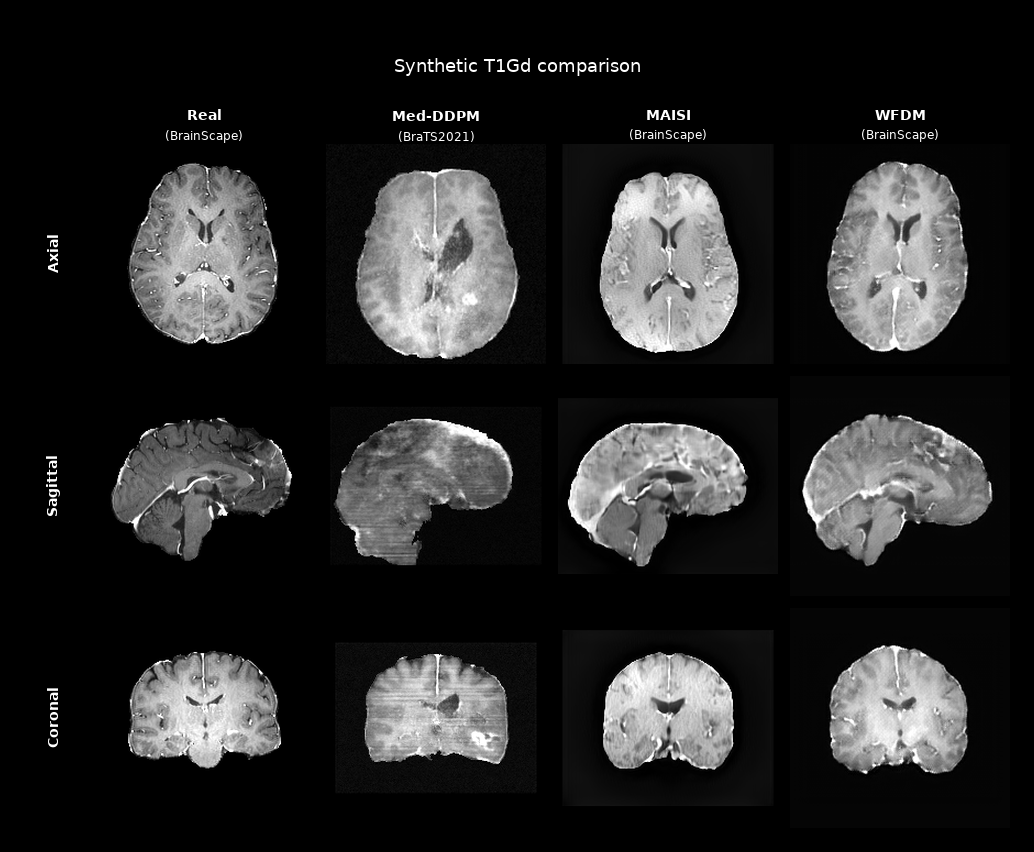}
  \caption{Qualitative comparison for T1Gd synthesis. The real BrainScape target image is compared with Med-DDPM (BraTS2021), MAISI (BrainScape), and WFDM (BrainScape) across axial, sagittal, and coronal views.}
  \label{fig:appendix-diff-t1ce}
\end{figure*}

\begin{figure*}[t]
  \centering
  \includegraphics[width=0.82\textwidth]{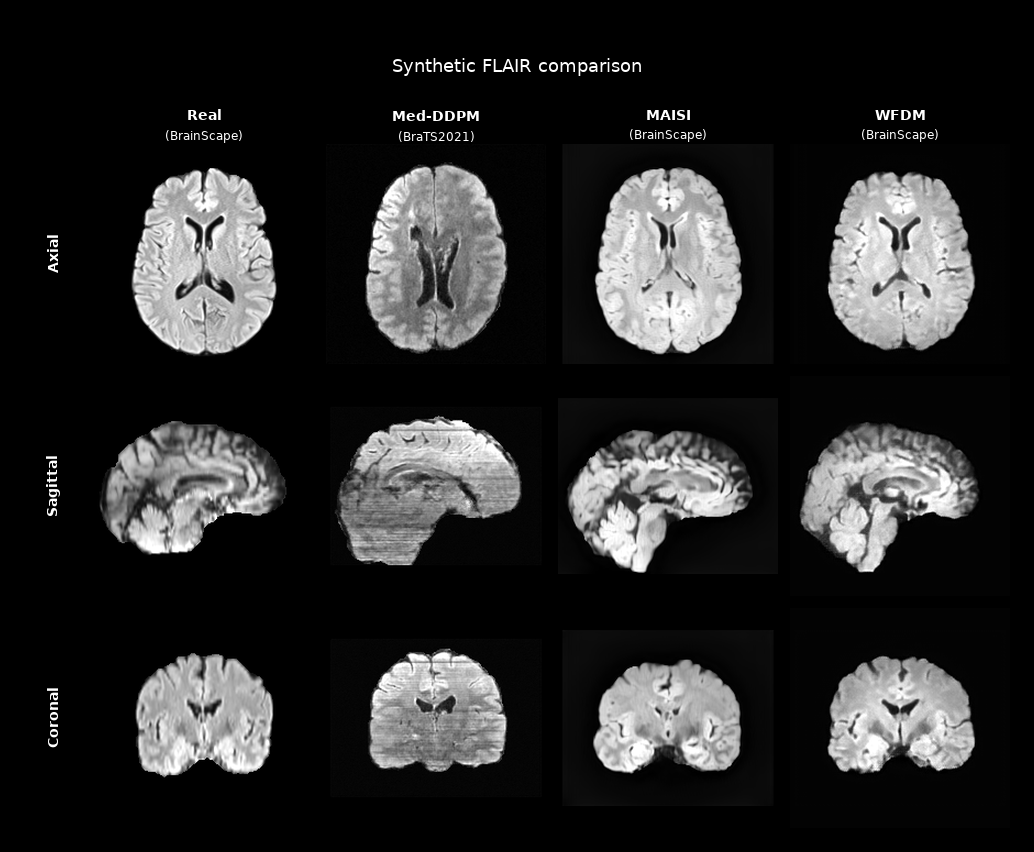}
  \caption{Qualitative comparison for FLAIR synthesis. The real BrainScape target image is compared with Med-DDPM (BraTS2021), MAISI (BrainScape), and WFDM (BrainScape) across axial, sagittal, and coronal views.}
  \label{fig:appendix-diff-flair}
\end{figure*}

This appendix extends the axial overview shown in the main paper by providing qualitative diffusion comparisons across axial, sagittal, and coronal views.
Figures~\ref{fig:appendix-diff-t1w}--\ref{fig:appendix-diff-flair} show comparisons across the evaluated synthetic MRI generators for each target modality, including T1w, T2w, T1Gd, and FLAIR. 
Each figure includes axial, sagittal, and coronal slices for a single target modality. 
Together, these figures complement the quantitative results in Appendix~B and make it easier to inspect anatomical sharpness, contrast fidelity, and consistency.
The qualitative comparisons show that the BrainScape-trained MAISI baseline and the proposed WFDM generally preserve anatomical structure and modality-specific contrast more faithfully than the external baseline models. 
The T1Gd case remains the most challenging qualitative setting, consistent with the modality-specific metrics reported in Appendix~B.

Figure~\ref{fig:appendix-diff-t1w} shows that the T1w modality is supported by all evaluated synthetic MRI generators. 
Among the external T1w-only generators, LDM is visually the strongest, but it remains less realistic than the two latent diffusion models trained on BrainScape when compared with the real MRI. 
HA-GAN and WDM appear noticeably blurrier and recover less fine cortical detail. 
Med-DDPM introduces visible striping artifacts and reduced volumetric consistency, particularly in the sagittal and coronal views. 
Overall, the two models trained on BrainScape produce the most anatomically coherent T1w syntheses, with WFDM providing the strongest qualitative T1w results, consistent with its best T1w FID and MMD scores in Appendix~B.

Figure~\ref{fig:appendix-diff-t2w} provides a qualitative comparison of T2w synthetic samples across the evaluated models.
Med-DDPM shows the strongest artifacts and poorest anatomical correspondence, particularly in the sagittal and coronal views, whereas the BrainScape-trained MAISI and WFDM models generate more realistic T2w contrast and preserve neuroanatomical structure more accurately. 
This visual pattern is consistent with the quantitative results in Appendix~B, where WFDM improves on the internal MAISI baseline for both FID and MMD in T2w synthesis.

Figure~\ref{fig:appendix-diff-t1ce} provides a qualitative comparison of T1Gd synthetic samples across the evaluated models. 
Med-DDPM produces the weakest result, with clear structural artifacts and poor enhancement fidelity. 
Both BrainScape-trained latent models are substantially better, but the MAISI internal baseline produces a cleaner and sharper image than WFDM across the three views. 
This mirrors the quantitative results in Appendix~B, where T1Gd is the only modality for which the MAISI baseline outperforms WFDM on FID and MMD.

Figure~\ref{fig:appendix-diff-flair} provides a qualitative comparison of FLAIR synthetic samples across the evaluated models.
This figure shows that both latent diffusion models trained on BrainScape generate plausible FLAIR anatomy, whereas Med-DDPM exhibits residual artifacts and weaker anatomical fidelity. 
WFDM is competitive with the MAISI baseline and produces sharper suppression patterns and cleaner anatomical structures than Med-DDPM. 
The difference between WFDM and MAISI is smaller for FLAIR than for T1Gd, which is consistent with the quantitative results in Appendix~B.

\FloatBarrier
\subsection{Appendix D: Modality-specific scheduler and sampling ablation results}

This appendix reports the full modality-specific scheduler and sampling ablation results for the proposed WFDM model family. 
Tables~\ref{tab:scheduler-t1w}--\ref{tab:scheduler-flair} provide FID, MMD, and MS-SSIM scores separately for T1w, T2w, T1Gd, and FLAIR.
These results complement the pooled scheduler analysis in the main text and show how sampler behavior varies across target modalities. 

Overall, the results for individual modalities are consistent with the pooled results, showing that RFlow provides the best overall balance between distributional realism, diversity, and sampling cost. 
RFlow is the strongest configuration across modalities while requiring only 50 inference steps, compared with 1000 steps for the DDPM variants.
Within the DDPM family, the cosine schedule generally performs better than the scaled-linear schedule on FID and MMD. 
Across all modalities, the scaled-linear DDPM configuration produces the lowest MS-SSIM between generated samples, indicating lower self-similarity among generated outputs, but this is coupled with substantially worse FID and MMD. 
Thus, lower MS-SSIM alone should not be interpreted as better generation quality when distributional alignment with real images deteriorates. 
These results support adopting RFlow with $v$-prediction and an $\ell_2$ objective as the default sampling configuration for BrainScape synthesis.

\begin{table*}[t]
\centering
\scriptsize
\begin{threeparttable}
\caption{Scheduler comparison for T1w.}
\label{tab:scheduler-t1w}
\setlength{\tabcolsep}{6pt}
\renewcommand{\arraystretch}{1.1}
\begin{tabular}{@{}lcccccc@{}}
\toprule
Scheduler / Sampler & Pred. type & Loss & Inference steps & FID $\downarrow$ & MMD $\downarrow$ & MS-SSIM $\downarrow$ \\
\midrule
DDPM (cosine $\beta$ schedule) & $v$ & $\ell_1$ & 1000 & 0.00359891 & 0.182717 & 0.790412 \\
DDPM (cosine $\beta$ schedule) & $v$ & $\ell_2$ & 1000 & 0.00242316 & 0.135932 & 0.772470 \\
DDPM (scaled-linear $\beta$ schedule) & $v$ & $\ell_2$ & 1000 & 0.05874300 & 0.362470 & 0.419145 \\
RFlow & $v$ & $\ell_2$ & 50 & 0.00159097 & 0.123779 & 0.786674 \\
\bottomrule
\end{tabular}
\end{threeparttable}
\end{table*}

\begin{table*}[t]
\centering
\scriptsize
\begin{threeparttable}
\caption{Scheduler comparison for T1Gd.}
\label{tab:scheduler-t1ce}
\setlength{\tabcolsep}{6pt}
\renewcommand{\arraystretch}{1.1}
\begin{tabular}{@{}lcccccc@{}}
\toprule
Scheduler / Sampler & Pred. type & Loss & Inference steps & FID $\downarrow$ & MMD $\downarrow$ & MS-SSIM $\downarrow$ \\
\midrule
DDPM (cosine $\beta$ schedule) & $v$ & $\ell_1$ & 1000 & 0.0247965 & 0.268550 & 0.746009 \\
DDPM (cosine $\beta$ schedule) & $v$ & $\ell_2$ & 1000 & 0.0242208 & 0.272274 & 0.740747 \\
DDPM (scaled-linear $\beta$ schedule) & $v$ & $\ell_2$ & 1000 & 0.0439559 & 0.298297 & 0.510649 \\
RFlow & $v$ & $\ell_2$ & 50 & 0.0241885 & 0.289871 & 0.740414 \\
\bottomrule
\end{tabular}
\end{threeparttable}
\end{table*}

\begin{table*}[t]
\centering
\scriptsize
\begin{threeparttable}
\caption{Scheduler comparison for T2w.}
\label{tab:scheduler-t2w}
\setlength{\tabcolsep}{6pt}
\renewcommand{\arraystretch}{1.1}
\begin{tabular}{@{}lcccccc@{}}
\toprule
Scheduler / Sampler & Pred. type & Loss & Inference steps & FID $\downarrow$ & MMD $\downarrow$ & MS-SSIM $\downarrow$ \\
\midrule
DDPM (cosine $\beta$ schedule) & $v$ & $\ell_1$ & 1000 & 0.00748287 & 0.226672 & 0.693967 \\
DDPM (cosine $\beta$ schedule) & $v$ & $\ell_2$ & 1000 & 0.01124230 & 0.249242 & 0.673993 \\
DDPM (scaled-linear $\beta$ schedule) & $v$ & $\ell_2$ & 1000 & 0.03167990 & 0.271586 & 0.512866 \\
RFlow & $v$ & $\ell_2$ & 50 & 0.00754393 & 0.218096 & 0.677926 \\
\bottomrule
\end{tabular}
\end{threeparttable}
\end{table*}

\begin{table*}[t]
\centering
\scriptsize
\begin{threeparttable}
\caption{Scheduler comparison for FLAIR.}
\label{tab:scheduler-flair}
\setlength{\tabcolsep}{6pt}
\renewcommand{\arraystretch}{1.1}
\begin{tabular}{@{}lcccccc@{}}
\toprule
Scheduler / Sampler & Pred. type & Loss & Inference steps & FID $\downarrow$ & MMD $\downarrow$ & MS-SSIM $\downarrow$ \\
\midrule
DDPM (cosine $\beta$ schedule) & $v$ & $\ell_1$ & 1000 & 0.00336125 & 0.116633 & 0.715843 \\
DDPM (cosine $\beta$ schedule) & $v$ & $\ell_2$ & 1000 & 0.00660881 & 0.147881 & 0.691909 \\
DDPM (scaled-linear $\beta$ schedule) & $v$ & $\ell_2$ & 1000 & 0.05816760 & 0.319046 & 0.403612 \\
RFlow & $v$ & $\ell_2$ & 50 & 0.00324606 & 0.116996 & 0.705553 \\
\bottomrule
\end{tabular}
\end{threeparttable}
\end{table*}

\FloatBarrier
\printbibliography

\end{document}